\documentclass[conference]{IEEEtran}
\usepackage{amsmath,amsfonts}
\usepackage{algorithmic}
\usepackage{algorithm}
\usepackage{textcomp}
\usepackage{url}
\usepackage{verbatim}
\usepackage{graphicx}
\usepackage{times}
\usepackage{epsfig}
\usepackage{amsmath}
\usepackage{amssymb}
\usepackage{graphicx}
\usepackage{booktabs}
\usepackage{multirow}
\usepackage[normalem]{ulem}
\usepackage{overpic}       
\usepackage{gensymb}
\usepackage{orcidlink}
\usepackage{authblk}

\useunder{\uline}{\ul}{}
\usepackage{etoolbox}
\AtBeginEnvironment{table}{}
\AtBeginEnvironment{table*}{}

\makeatother

\graphicspath{{./image/}{./}}

\begin{document}

\title{EchoSR: Efficient Context Harnessing for Lightweight Image Super-Resolution}

\author[1]{Hanli Zhao}
\author[1]{Binhao Wang}
\author[1]{Shihao Zhao}
\author[2]{Tao Wang}
\author[3]{Kaihao Zhang}
\author[1,4]{Wanglong Lu$^{*}$}

\affil[1]{College of Computer Science and Artificial Intelligence, Wenzhou University, Wenzhou 325000, China}
\affil[2]{vivo BlueImage Lab, vivo Mobile Communication Co., Ltd, Shanghai 200100, China}
\affil[3]{College of Engineering and Computer Science, Australian National University, Canberra, Australia}
\affil[4]{The AI/Analytics Team, Nasdaq, St. John's, Canada}
\affil[$^{*}$]{Corresponding author: lwlxhl@gmail.com}

\maketitle
\begin{abstract}
Image super-resolution (SR) aims to reconstruct high-quality, high-resolution (HR) images from low-resolution (LR) inputs and plays a critical role in various downstream applications.
Despite recent advancements, balancing reconstruction fidelity and computational efficiency remains a fundamental challenge, particularly in resource-constrained scenarios.
While existing lightweight methods attempt to expand receptive fields, many of them either incur substantial computational overhead, naively scale up kernel sizes, or lack mechanisms for coherent multi-scale integration, limiting their overall effectiveness and scalability.
To address these limitations, we propose EchoSR, an efficient context-harnessing framework for lightweight image super-resolution, which unifies multi-scale receptive field modeling and hierarchical context fusion.
EchoSR decouples feature learning into disentangled local, multi-scale, and global modeling stages through an efficient context-harnessing strategy, and further promotes seamless cross-scale integration via a cross-scale overlapping fusion mechanism.
Extensive experiments have shown that EchoSR consistently outperforms state-of-the-art lightweight super-resolution methods across multiple benchmarks, while also achieving a faster speed $(\sim 2\times)$. The source code is available at \url{https://github.com/funnyWang-Echoes/EchoSR}.
\end{abstract}



\section{Introduction}\label{Introduction}
Image super-resolution (SR), as a fundamental task in computer vision, aims to reconstruct high-quality and high-resolution (HR) images from low-resolution (LR) inputs. 
With the rapid development of ultra-high-definition display devices and mobile applications, SR technology has demonstrated significant value in various fields, including medical imaging~\cite{SSEFusion,Entropy-aware}, video surveillance~\cite{Fusion-driven}, satellite remote sensing~\cite{LTFormer,MDGF-CD}, and image editing~\cite{FACEMUG,LU2025111312}. 
However, achieving an optimal balance between reconstruction quality and computational efficiency remains a critical challenge in resource-constrained scenarios.
\begin{figure}[!htbp]
\centering
\includegraphics[width=1\columnwidth]{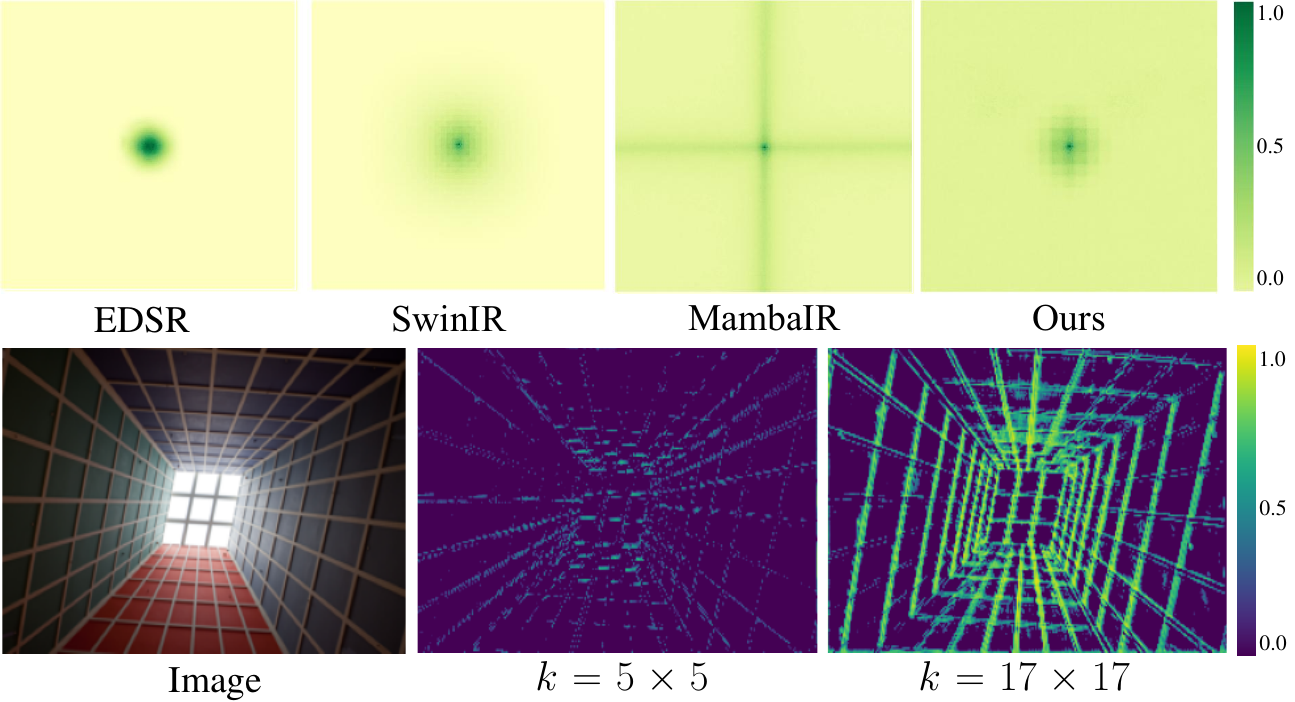}
\caption{Visualization of the effective receptive field (ERF) (top) and the feature maps of different kernel sizes (bottom). The colors represent the receptive field distribution and activation intensity, respectively.}
\label{fig:firstFig}
\end{figure}

A core difficulty in lightweight SR arises from the need to balance receptive field expansion with computational cost. 
Convolutional neural networks (CNNs), while effective at capturing local patterns, often struggle to model long-range dependencies without substantially increasing model depth~\cite{DDistill-SR,EDSR,IMDN}. 
Conversely, some Transformer-based approaches excel at global context modeling but may suffer from quadratic computational complexity~\cite{swinIR,DAT,swinNG,SRFormer}.
Recently, state space models (SSMs), exemplified by Mamba~\cite{gu2024mamba}, have emerged as efficient alternatives to self-attention, offering linear computational complexity. However, their inherent sequential nature poses challenges for high-fidelity SR~\cite{MambaIR,MaIR}. As MaIR~\cite{MaIR} points out, flattening 2D images into 1D sequences disrupts the locality and continuity intrinsic to visual data, necessitating additional designs to compensate for this loss. Furthermore, pixel-wise aggregation overlooks distinct contexts across different scanning paths, thereby hindering the exploitation of critical multi-scale spatial correlations essential for reconstructing fine details.

As shown in Fig.~\ref{fig:firstFig} (top), we visualize the effective receptive fields (ERF) of representative methods from CNN, Transformer, and Mamba architectures. EDSR~\cite{EDSR} focuses on local receptive field, while SwinIR~\cite{swinIR}, MambaIR~\cite{MambaIR} exhibit large receptive fields, though they still have limited coverage in areas farther from the center. Additionally, their computational efficiency remains a challenge, especially when deployed in resource-constrained environments.
We thus explore a novel approach to balance computational efficiency with large receptive fields, ultimately enhancing super-resolution performance.

To address this challenge, recent efforts have explored enlarging convolutional receptive fields using large-kernel designs as a trade-off solution~\cite{LKASR,MAN}. However, these methods primarily rely on sparse sampling via dilated convolutions with small kernels (e.g., $3 \times 3$ or $5 \times 5$), which often suffer from the gridding effect~\cite{RepLKNet} and lack explicit mechanisms for the deep fusion of information across different scales.
To better capture both local details and global structural information, we further investigate the impact of multi-scale kernels for feature extraction. 
As shown in Fig.~\ref{fig:firstFig} (bottom), larger kernels can effectively capture global and structural patterns, while smaller kernels excel at capturing local details. These findings suggest that direct, ultra-large kernels (up to $17 \times 17$) can serve as dense structural anchors. Combined with efficient depthwise implementation and a multi-branch channel splitting strategy, such a design offers a promising path to balance expansive receptive fields with low computational overhead. To complement the multi-scale structural expansion, we introduce a lightweight global context branch. Recognizing that pooling operations can effectively capture broad spatial features with negligible computational overhead~\cite{poolFormer}, this branch acts as an efficient global feature modulator, providing essential structural priors without the parameter burden of standard self-attention.

To this end, we propose EchoSR, an \textbf{E}fficient \textbf{c}ontext-\textbf{h}arnessing framew\textbf{o}rk for lightweight image \textbf{S}uper-\textbf{R}esolution, where hierarchical information fusion serves as the core architectural philosophy. EchoSR decouples the learning process into disentangled local, multi-scale, and global modeling stages. Recognizing the complementary nature of these decoupled features, we introduce a streamlined yet effective concatenation strategy. Specifically, the features from different perspectives are integrated via channel-wise concatenation followed by a Feed-Forward Network (FFN) that acts as a deep feature mixer to facilitate dynamic coordination and recalibration across branches. Second, effective information propagation across scales is essential for preserving contextual continuity; thus, we introduce a cross-scale overlapping fusion mechanism within EchoSR that promotes gradual transitions between scales, ensuring the harmonious integration of fine-grained details and high-level structures throughout the feature learning process.

\begin{figure}[t]
\centering
\includegraphics[width=1\linewidth]{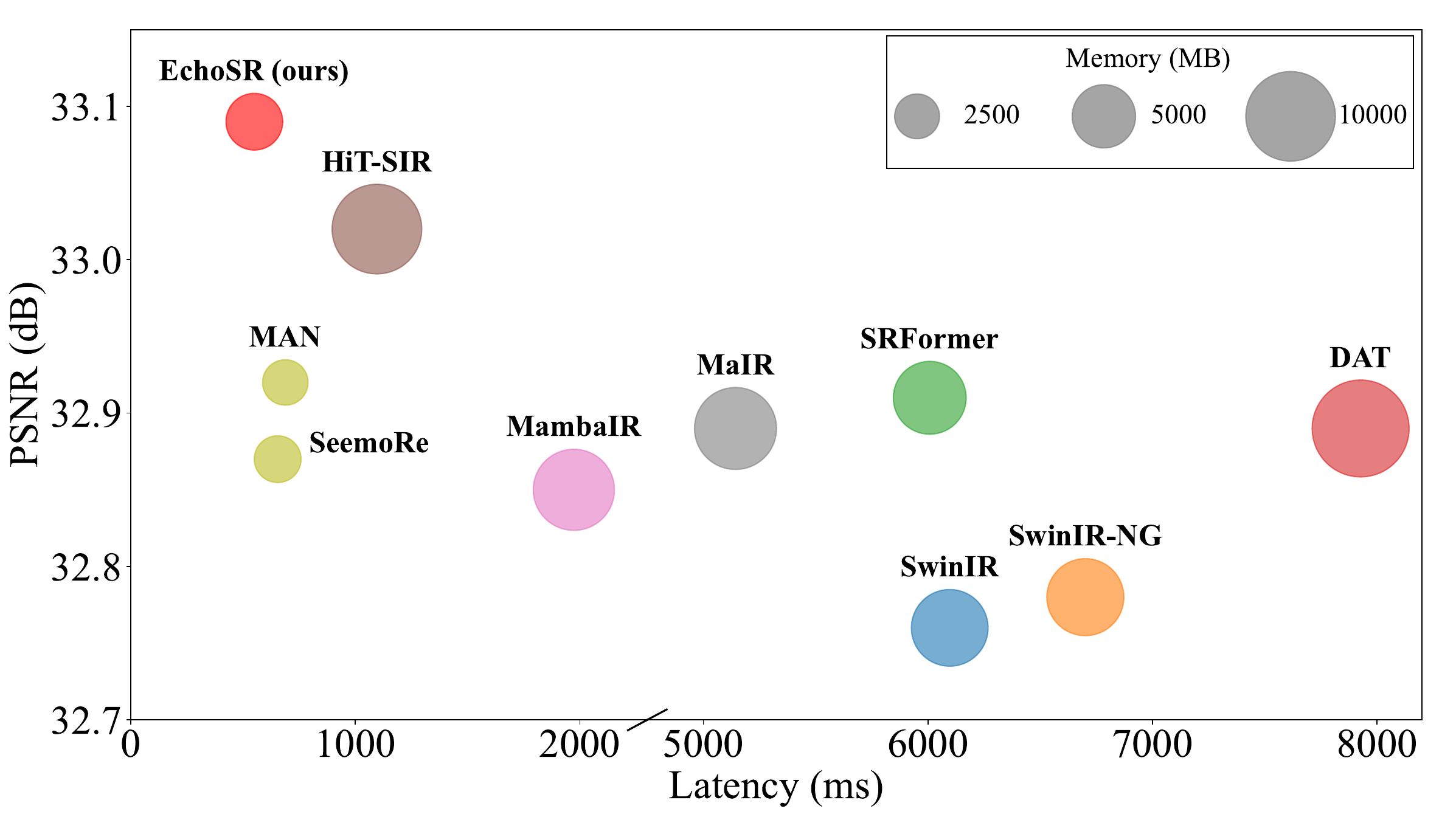}
\caption{Comparisons on the Urban100 test set at $\times2$ scale with input resolution of $1024 \times 1024$. The area of each circle indicates peak memory usage during inference. EchoSR demonstrates the best balance among performance, memory consumption, and inference latency.}
\label{psnrvs}
\end{figure}

Building on these two principles, we first introduce an efficient context-harnessing strategy that unfolds the learning process into disentangled local, multi-scale, and global modeling stages:
(1) Local aggregation (LA) focuses on enhancing fine-grained representations via balanced channel expansion and group convolutions, ensuring that local details are preserved with efficiency;
(2) Multi-scale receptive field expansion (MRFE) captures spatial structures at varying scales through parallelized large-kernel depthwise convolutions, enabling the model to flexibly adjust its receptive field according to feature complexity;
(3) Global context prior (GCP) provides structural priors by employing lightweight downsampling-convolution-upsampling operations.
Secondly, we introduce a Cross-scale Overlapping Fusion Block (COFB) to facilitate systematic cross-scale fusion. By constructing overlapping receptive fields through cascaded large-kernel operations, COFB achieves spatial structural rectification by aligning features from different spatial hierarchies. This mechanism ensures a gradual and coherent transition between scales, allowing fine-grained details and macro-structures to be harmoniously integrated and fused throughout the feature learning process.

By orchestrating hierarchical feature extraction and seamless cross-scale fusion under a unified framework, EchoSR effectively coordinates representational dependencies across spatial hierarchies.
Extensive experiments demonstrate that EchoSR delivers high-fidelity super-resolution with significantly reduced computational cost, as illustrated in Fig.~\ref{psnrvs}.
The contributions of this paper can be summarized as follows:
\begin{itemize}
\item We propose EchoSR, an efficient context-harnessing framework that unifies fine-grained detail enhancement and global structure fusion through hierarchical and overlapping context modeling strategies.
\item We design an efficient context-harnessing mechanism, comprising local aggregation, multi-scale receptive field expansion, and global context prior, to achieve dynamic adaptation to spatial feature diversity.
\item We introduce a contextual overlapping fusion strategy to ensure seamless continuity across scales, promoting comprehensive feature integration and mitigating multi-scale discontinuities.
\item Extensive experiments across multiple benchmarks demonstrate that EchoSR consistently surpasses state-of-the-art (SOTA) lightweight SR methods, providing an efficient and scalable solution for practical super-resolution applications.
\end{itemize}

\section{Related Work}
\subsection{Image Super-Resolution}
CNNs were first used to solve the image super-resolution problem in deep learning. SRCNN pioneered the use of a simple 3-layer convolutional network to map between HR and LR images, surpassing traditional interpolation methods. Subsequent CNN-based approaches have refined and extended SRCNN. VDSR~\cite{VDSR} and EDSR~\cite{EDSR} leveraged residual learning to construct deeper networks. However, these methods inherently rely on the inductive bias of CNNs, focusing on learning local features. 
Recently, some works~\cite{LKASR,MAN,SeeMore} have turned to how to enhance the receptive field of CNNs with large kernels.

Self-attention mechanisms have been widely adopted in vision tasks to capture long-range dependencies. SwinIR~\cite{swinIR} designed a shifted window self-attention mechanism, becoming a representative work in super-resolution. However, due to the quadratic complexity of self-attention, the limited window size restricts further long-range dependency modeling. ELAN~\cite{ELAN} reduces computational complexity by decreasing feature dimensionality, limiting the network's potential.  SRFormer~\cite{SRFormer} and DAT~\cite{DAT} increase the window size, which improves performance but also introduces additional computational burden. Recently, HiT-SIR~\cite{HiT-SR} achieved efficient super-resolution by introducing a hierarchical design and employing different window sizes. Nevertheless, Transformer-based methods suffer from high computational cost and poor scalability for high-resolution inputs.

Inspired by structured state-space sequence models, Mamba~\cite{Mamba} provides a novel and efficient mechanism for feature sequence processing. In contrast to self-attention, the trainable parameters within Mamba are determined by the dimensionality of individual features within the sequence rather than the sequence length itself. It leverages a selection mechanism to effectively process discrete and information-rich data. MambaIR~\cite{MambaIR} pioneered the integration of the Mamba architecture into image reconstruction, achieving promising results. Meanwhile, MaIR~\cite{MaIR} explores using S-shaped nested scanning. Although SSMs offer linear theoretical complexity, their practical inference efficiency is currently constrained by the early stage of operator optimization.
Furthermore, the inherently sequential nature of SSMs may underutilize critical multi-scale and local spatial correlations, which are essential for high-fidelity SR.

Our work deeply studies the architecture of large-kernel convolution. By employing efficient context-harnessing and contextual overlapping fusion strategies in large-kernel convolution, we achieve high performance and efficiency.

\begin{table}[]
\centering
\caption{Comparison of EchoSR with other CNN-based methods. We partition methods into general-purpose backbones and SR-specific networks. Given that an initial $3\times3$ convolution is universally employed for preliminary local feature extraction across all architectures, our comparison centers specifically on the distinctive design of the \textbf{backbone block}. We evaluate the receptive field scope (Local, Multi-scale, Global) and feature fusion strategies employed within these blocks. Local denotes an explicit local modeling mechanism.} Multi-scale refers to architectures employing mixed kernel sizes. Global encompasses methods utilizing large kernels (simulating global receptive fields), spatial gate attention mechanisms, or pooling strategies to capture broad context. FFN denotes Feed-Forward Networks, encompassing standard MLPs and their optimized variants.
\label{tab:diff}
\resizebox{1\columnwidth}{!}{%
\begin{tabular}{@{}ccccc@{}}
\toprule
\textbf{Methods} & \textbf{Local} & \textbf{Multi-scale} & \textbf{Global} & \textbf{Fusion strategy} \\ \midrule
RepLKNet~\cite{RepLKNet}        &   &           & \textbf{\checkmark} & FFN               \\
VAN~\cite{VAN}            &   &           & \textbf{\checkmark} & Gate+FFN          \\
SLaK~\cite{SLaK}            &   &           & \textbf{\checkmark} & FFN               \\
InceptionNeXt~\cite{InceptionNeXt}   &   & \textbf{\checkmark}         &   & FFN               \\
UniRepLKNet~\cite{UniRepLKNet}     &   & \textbf{\checkmark}         & \textbf{\checkmark} & FFN               \\ \midrule
IMDN~\cite{IMDN}            & \textbf{\checkmark} &           &   & Channel attention \\
LKASR~\cite{LKASR}           &   &           & \textbf{\checkmark} & FFN               \\
SeemoRe~\cite{SeeMore}         & \textbf{\checkmark} &           & \textbf{\checkmark} & FFN               \\
MAN~\cite{MAN}             &   & \textbf{\checkmark} &   & FFN          \\
CFSR~\cite{Wu_cfsr}            & \textbf{\checkmark}  &           & \textbf{\checkmark} & Gate+FFN               \\
EchoSR        & \textbf{\checkmark} & \textbf{\checkmark}  & \textbf{\checkmark} & FFN+COFB          \\ \bottomrule
\end{tabular}%
}
\end{table}

\subsection{Large Kernel in Convolutions}

ConvNeXt~\cite{ConvNet} introduced $7 \times 7$ depthwise convolutions, surpassing Swin Transformer in high-level vision tasks. This sparked renewed interest in large-kernel convolutions. RepLKNet~\cite{RepLKNet} achieved a $31 \times 31$ convolutional receptive field. VAN~\cite{VAN} stacked depthwise and depthwise dilated convolutions, and mimicked the attention mechanism to implement large kernel attention (LKA) with a $21 \times 21$ receptive field, providing a novel paradigm. Further analyzing the potential of large kernels, SLaK~\cite{SLaK} demonstrated that a kernel of sufficient size can function as a surrogate for global attention. By leveraging large-kernel decomposition, SLaK realized an equivalent $51 \times 51$ convolution. More recently, InceptionNeXt~\cite{InceptionNeXt} and UniRepLKNet~\cite{UniRepLKNet} have further substantiated the effectiveness of large-kernel designs.

In the context of super-resolution, existing large-kernel frameworks primarily rely on dilated convolutions such as LKASR~\cite{LKASR} and MAN \cite{MAN}, or specific geometries to expand the receptive field like SeemoRe~\cite{SeeMore} employs strip convolutions to complement its local and global experts. While efficient, these sparse sampling strategies are susceptible to the gridding effect because they typically utilize smaller $3 \times 3$ or $5 \times 5$ kernels with high dilation rates~\cite{RepLKNet}, which can disrupt the fine-grained spatial correlations essential for texture reconstruction. CFSR~\cite{Wu_cfsr} explores a ConvFormer-based approach where a $9\times9$ depthwise convolution serves as a gated feature mixer to replace self-attention for efficient global contextual modeling. CFSR also introduces a re-parameterization strategy to achieve local feature preservation.

Table~\ref{tab:diff} summarizes the differences between our method and other methods. EchoSR represents a significant mechanistic departure from these paradigms by emphasizing dense spatial sampling and decoupled context harnessing. Specifically, unlike dilation-based methods, we employ direct depthwise convolutions up to $17 \times 17$ to ensure a continuous receptive field that preserves spatial integrity and representation density. Furthermore, EchoSR introduces a decoupled learning paradigm that disentangles feature extraction into localized aggregation, multi-scale structural expansion, and efficient global context modeling. Unlike approaches that rely on ultra-large kernel decomposition~\cite{RepLKNet,SLaK,UniRepLKNet}, or dilated convolutions to capture global information~\cite{LKASR,MAN}, we employ a pooling strategy~\cite{poolFormer} to supplement global features efficiently. This decoupled architecture ensures that large kernels are not merely monolithic feature mixers but are synergistically integrated with diverse spatial priors. This architectural novelty is further enhanced by the COFB module, which moves beyond simple feature concatenation to implement a small-to-large cascaded rectification process. This strategy effectively mitigates structural response attenuation and ensures that the aggregated long-range dependencies are spatially calibrated, thereby achieving an optimal balance between global structural coherence and local texture fidelity.

\section{Method}

\subsection{Overview}

\begin{figure*}[!ht]
\centering
\includegraphics[width=1\textwidth]{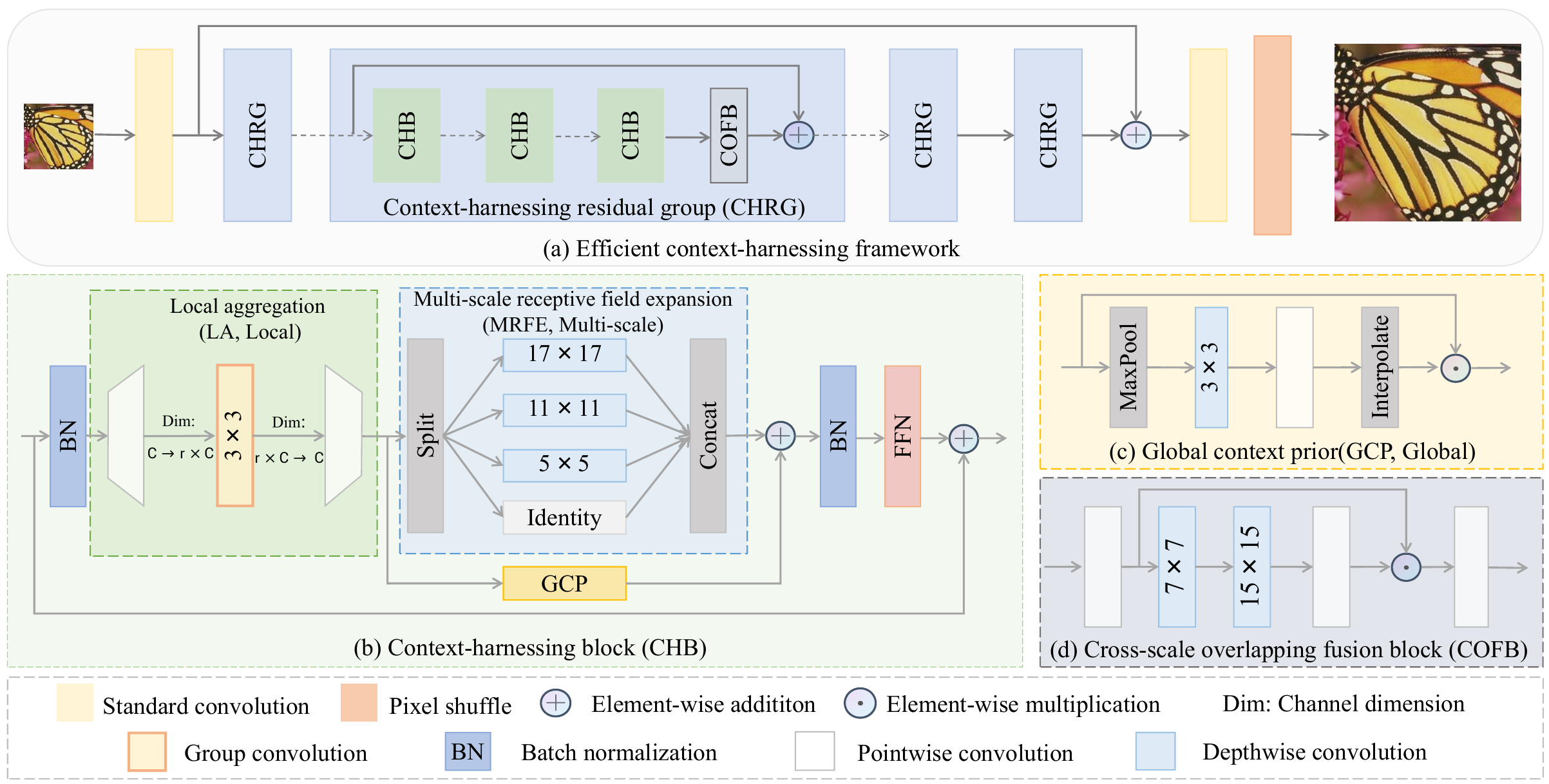}
\caption{Overview of our EchoSR architecture for lightweight image super-resolution. CHB extracts local, multi-scale, and global features in parallel, while COFB facilitates effective cross-scale feature fusion. The kernel size $k$ is specified for convolutions.}
\label{model1}
\end{figure*}

Our EchoSR decouples the learning process into three complementary perspectives: local, multi-scale, and global. 
Based on this principle, we design a Context-Harnessing Block (CHB) that: (1) fully utilizes local information through Local Aggregation (LA), (2) captures spatial structures at varying scales via Multi-Scale Receptive Field Expansion (MRFE), and (3) leverages long-range dependencies through Global Context Prior (GCP).
To promote coherent integration across scales, we introduce a Cross-Scale Overlapping Fusion Block (COFB) that constructs overlapping receptive fields through cascaded large-kernel operations. 
For enhanced high-frequency detail preservation and gradient flow, we establish Context-Harnessing Residual Groups (CHRGs) using CHB, COFB, and residual connections between shallow features and deep feature outputs.
These components are integrated into our Efficient Context-Harnessing Network.

As shown in Fig.~\ref{model1} (a), given an input low-resolution image $I_{LR} \in \mathbb{R}^{H \times W \times 3}$, EchoSR first applies a $3 \times 3$ convolution to extract shallow features $X \in \mathbb{R}^{H \times W \times C}$. These features are further refined through $N$ stacked CHRGs to model complementary contextual cues at local, multi-scale, and global scales.
Finally, a $3 \times 3$ convolution followed by a pixel-shuffle module generates the super-resolution output $I_{SR}$.

Specifically, the entire process can be described in the following three stages:

\textbf{1) Shallow feature extraction.} Given an input low-resolution image $I_{LR} \in \mathbb{R}^{H \times W \times 3}$, EchoSR first applies a $3 \times 3$ standard convolution layer $H_{sf}(\cdot)$ to map the image into the feature space:
\begin{equation}
    X_{0} = H_{sf}(I_{LR}),
\end{equation}
where $X_{0} \in \mathbb{R}^{H \times W \times C}$ represents the shallow features, and $C$ denotes the channel dimension.

\textbf{2) Deep feature extraction.} The shallow features are further refined through $N$ cascaded CHRGs. For the $i$-th group, the operation can be formulated as:
\begin{equation}
    X_{i} = \operatorname{COFB}(\operatorname{CHB}_{L}(\dots \operatorname{CHB}_{1}(X_{i-1}))) + X_{i-1}, \quad i = 1, 2, \dots, N
\end{equation}
where $\operatorname{CHB}(\cdot)$ and $\operatorname{COFB}(\cdot)$ denote the Context-Harnessing Block and Cross-scale Overlapping Fusion Block, respectively. Within each CHRG, a series of $L$ blocks of CHB extract complementary contextual cues in parallel, which are then integrated through a COFB module to ensure cross-scale spatial consistency. A group-level residual connection is established to stabilize the gradient flow.

\textbf{3) Image reconstruction.} After $N$ groups of refinement, we obtain the deep features $X_{N}$. EchoSR adopts a global residual learning strategy to generate the final super-resolution output:
\begin{equation}
    I_{SR} = H_{rec}(X_{0} + X_{N}),
\end{equation}
where $H_{rec}(\cdot)$ denotes the reconstruction module, which consists of a $3 \times 3$ convolution followed by a pixel-shuffle layer to upsample the features to the target resolution $I_{SR} \in \mathbb{R}^{rH \times rW \times 3}$, with $r$ being the upscaling factor.

\subsection{Context-Harnessing Block}

Effective feature context harnessing is essential for neural networks to focus on meaningful patterns while suppressing noise during feature extraction and reasoning. This relies on modeling rich and relevant relationships across spatial regions. 
Our unified framework (EchoSR) integrates all three perspectives (local, multi-scale, and global) into a hierarchical learning process.
This enables our method to achieve a balanced trade-off between expressive capacity and computational efficiency, while fully leveraging diverse contextual cues, and ultimately achieve high-quality image super-resolution.

Given the importance of both multi-scale coordination and global context for high-fidelity super-resolution, we investigate existing context expansion modeling strategies in CNNs. These methods predominantly fall into two streams: (1) large-kernel based approaches, which utilize large kernels~\cite{ConvNet,SLaK} or gate attention mechanisms~\cite{VAN,Wu_cfsr,LKASR} to expand the receptive field and emulate the global representations; (2) multi-scale strategies~\cite{mixConv,InceptionNeXt,MAN}, which build broader context by mixing convolutions with varying kernel sizes. However, by focusing heavily on expanding the receptive field, these approaches may inadvertently compromise fine-grained local details that are crucial for super-resolution. Furthermore, they rarely incorporate a dedicated branch to capture image-wide statistics alongside convolution-based features. To address this, we propose to explicitly disentangle feature extraction into three complementary branches: local aggregation for fine details, multi-scale expansion for structural patterns, and global context prior for image-wide information.

As shown in Fig.~\ref{model1} (b), our context-harnessing block (CHB) includes a local aggregation (LA), a multi-scale receptive field expansion (MRFE), and a global context prior (GCP). Then, a feed-forward network (FFN) is further applied to refine the fused feature, incorporating residual connections.
Given an input feature $X \in \mathbb{R}^{H \times W \times C}$, the CHB is described as follows:
\begin{align}
    & X'= \operatorname{LA}(\operatorname{BN}(X)), \nonumber \\
    & X''= \operatorname{MRFE}(X') + \lambda \operatorname{GCP}(X'), \nonumber \\
    & X_{out}= \operatorname{FFN}(\operatorname{BN}(X'')) + X,
    \label{eq:CHB}
\end{align}
where $\operatorname{BN}(\cdot)$ represents Batch Normalization and $\lambda$ is a learnable scaling coefficient.

\textbf{Local aggregation module.} Local features are an indispensable component in SR tasks. Early SR methods~\cite{SRCNN,EDSR}, achieved remarkable performance by relying on the local inductive bias of convolutional features. We propose a local aggregation (LA), which employs two pointwise convolutional layers with an intermediate group convolution layer to aggregate and enhance the capture of local dependencies.
For a given input feature $X \in \mathbb{R}^{H \times W \times C}$, the entire process of the LA can be described as follows:
\begin{equation}
\operatorname{LA}(X) = \operatorname{PWC}^{r \times c \rightarrow c}(\operatorname{GC}(\operatorname{PWC}^{c \rightarrow r \times c}(X)))),
\label{eq:LA}
\end{equation}
where $\operatorname{GC}(\cdot)$ denotes group convolution and $\operatorname{PWC}(\cdot)$ denotes pointwise convolution.
In the LA, the number of channels in the input feature map is first expanded through a dimension expansion scheme, using pointwise convolution to scale the original channel count from $c$ to $r \times c$, where $r$ is a hyperparameter, and we set $r=1.5$ in this paper. This expansion enhances the network's learning capacity, enabling it to capture richer feature information. Subsequently, the feature map processed by the $3 \times 3$ convolutional layer undergoes a dimension compression scheme, where another pointwise convolution is applied to restore the channel count to $c$. 
Additionally, we employ a group convolution strategy in the LA, ensuring a good balance between efficiency and computational costs. We divide the input feature map into groups of 6 channels each and independently apply convolution operations to each group.

\textbf{Multi-scale receptive field
expansion.} To effectively capture multi-scale information and spatial dependencies, we design a Multi-Scale Receptive Field Expansion (MRFE) module. This module employs convolutional kernels of different sizes in parallel, enabling the network to capture features across multiple spatial ranges and significantly expand the receptive field.

Specifically, we split the input features evenly into four groups along the channel dimension. Each group is processed with a different kernel $k \times k$, including $5 \times 5$, $11 \times 11$, $17 \times 17$, and an identity mapping, respectively. The use of multi-scale kernels improves the model’s ability to recognize objects of varying sizes, while large kernels enhance long-range context modeling. The identity mapping branch preserves original features and reduces unnecessary computation.

Given an input feature $X' \in \mathbb{R}^{H \times W \times C}$ from LA, MRFE is described as follows:
\begin{align}
    &X_{id}, X_1, X_2, X_3 = \operatorname{Split}(X'), \nonumber \\
    &X_{id}' = \operatorname{Identity}(X_{id}),
    X_1' = \operatorname{DWC}_{5 \times 5}(X_1),\nonumber \\
    &X_2' = \operatorname{DWC}_{11 \times 11}(X_2),
    X_3' = \operatorname{DWC}_{17 \times 17}(X_3), \nonumber \\
    &\operatorname{MRFE}(X') = \operatorname{Concat}(X_{id}', X_1', X_2', X_3'),
    \label{eq:MRFE}
\end{align}
where $\operatorname{Split}(\cdot)$ represents the channel splitting operation, $\operatorname{DWC}_{k \times k}(\cdot)$ represents depthwise convolution with kernel size $k \times k$,  $\operatorname{Identity}(\cdot)$ represents the identity mapping, and $\operatorname{Concat(\cdot)}$ means concatenate operation.

\textbf{Global context prior.} The empirical success of PoolFormer~\cite{poolFormer} establishes that pooling mechanisms can effectively serve as a lightweight surrogate for self-attention in extracting global features. Drawing upon this insight, we further incorporate a Global Context Prior (GCP) module that captures high-level contextual information by operating on downsampled feature maps. This design complements MRFE by providing a broader receptive field at a lower computational cost, enabling the model to better interpret complex spatial relationships across the image.

Specifically, as shown in Fig.~\ref{model1} (c), we first apply max pooling to downsample the input features by a factor of 8, followed by a $3 \times 3$ depthwise convolution and a pointwise convolution. Then, the feature map is upsampled to the original resolution using bilinear interpolation to generate a global representation attention map. 
This attention map is multiplied with the input features using element-wise multiplication to obtain the final global feature map. The process can be formally expressed as:
\begin{align}
   &Att =\operatorname{UP}(\operatorname{PWC}(\operatorname{DWC}_{3\times3}(\operatorname{MaxPool}(X')))) \nonumber,\\
   &\operatorname{GCP}(X') = Att \odot X',
    \label{eq:GCP}
\end{align}
where $\operatorname{UP}(\cdot)$ denotes bilinear interpolation for upsampling, $\operatorname{MaxPool}(\cdot)$ represents the max pooling operation and $\odot$ means element-wise multiplication.

\subsection{Cross-Scale Overlapping Fusion Block}
In SR methods like SwinIR~\cite{swinIR}, a $3 \times 3$ convolution is typically added at the end of each group to strengthen local feature extraction in self-attention frameworks. However, in CNN-based SR models, the limited receptive field of convolutional kernels makes it challenging to effectively combine local and global information. To tackle this, we introduce the Cross-Scale Overlapping Fusion Block (COFB), which enhances feature interaction by designing overlapping receptive fields. This allows the network to better capture fine local details while incorporating broader contextual cues.

As shown in Fig.~\ref{model1} (d), COFB begins with a pointwise convolution to transform the input features, followed by two cascaded depthwise large-kernel convolutional layers with kernel sizes $k=7$ and $k=15$, respectively, followed by a pointwise convolution. Through this overlapping design, the two convolutional layers can cover receptive fields of different scales, effectively capturing both fine-grained local details and broader global contextual information. Finally, COFB performs element-wise multiplication between the input features and the output of the cascaded depthwise large-kernel convolutions, followed by another pointwise convolution to modulate and enhance the feature representation, leading to improved feature quality.

Given the output feature $X_{out} \in \mathbb{R}^{H \times W \times C}$ from CHB, we get aggregated output feature $Y$ after COFB:
\begin{align}
&\hat{Y} = \operatorname{PWC}(X_{out}),\nonumber \\
&\bar{Y}=\operatorname{PWC}(\operatorname{DWC}_{15 \times 15}(\operatorname{DWC}_{7 \times 7}(\hat{Y}))), \nonumber \\
&Y = \operatorname{PWC}(\bar{Y} \odot \hat{Y}).
\label{eq:COFB}
\end{align}

\subsection{Loss Functions}

To train our SR model, we employ the following combined loss function:
\begin{equation}
\mathcal{L}_{total} = \mathcal{L}_{pixel} + \alpha \mathcal{L}_{freq}, \label{eq:total_loss}
\end{equation}
where $\mathcal{L}_{pixel}$ is the pixel-wise loss function, which measures the difference between the output image of the model and the ground truth image at the pixel level. We adopt the $L_1$ norm for this purpose:
\begin{equation}
\mathcal{L}_{pixel} = \|I_{SR} - I_{GT}\|_1 ,\label{eq:pixel_loss}
\end{equation}
where $I_{SR}$ represents the super-resolution image from the model, and $I_{GT}$ denotes the corresponding ground truth image.

$\mathcal{L}_{freq}$ is the frequency domain loss function, designed to constrain the similarity between the model's output image and the ground truth image in the frequency domain. As pointed out in~\cite{ShuffleMixer}, constraining the frequency domain can help the model better learn high-frequency information of images. We apply the Fast Fourier Transform (FFT) to the images and compute the $L_1$ norm of their difference in the frequency domain:
\begin{equation}
\mathcal{L}_{freq} = \|\mathcal{F}(I_{SR}) - \mathcal{F}(I_{GT})\|_1 ,\label{eq:freq_loss}
\end{equation}
where $\mathcal{F}$ represents the Fast Fourier Transform. In this paper, we empirically set $\alpha$ to 0.1~\cite{ShuffleMixer,SeeMore}.

\begin{table}[htbp]
\centering
\caption{Detailed implementation and training configurations for EchoSR and EchoSR-lite.}
\label{tab:training_details}
\resizebox{1\columnwidth}{!}{
\begin{tabular}{lcc}
\toprule
\textbf{Parameter}                  & \textbf{EchoSR}      & \textbf{EchoSR-lite} \\
\midrule
\multicolumn{3}{l}{\textit{Architecture Configurations}} \\
Num. CHRGs                        & \multicolumn{2}{c}{4}        \\
CHBs Distribution                 & [5, 5, 5, 5] & [2, 3, 2, 3] \\
Channel Dimension ($C$)           & 60           & 36           \\
Expansion Ratio $r$ in LA         & \multicolumn{2}{c}{1.5}      \\
Kernel Sizes in MRFE              & \multicolumn{2}{c}{[Identity, 5, 11, 17]} \\
Kernel Sizes in COFB              & \multicolumn{2}{c}{7, 15}    \\
GCP Scaling Coefficient $\lambda$ & \multicolumn{2}{c}{Initialized to 0.1} \\
Padding Mode                      & \multicolumn{2}{c}{Zero Padding} \\
\midrule
\multicolumn{3}{l}{\textit{Training Settings}} \\
Training Dataset                  & DIV2K / DF2K   & DF2K         \\
LR Patch Size                     & \multicolumn{2}{c}{$72 \times 72$} \\
Data Augmentation                 & \multicolumn{2}{c}{Horizontal flip, Random rotation ($0^\circ,90^\circ, 180^\circ, 270^\circ$)} \\
Optimizer                         & \multicolumn{2}{c}{Adam ($\beta_1=0.9, \beta_2=0.99$)}     \\
Batch Size                        & \multicolumn{2}{c}{32}       \\
Total Iterations                  & \multicolumn{2}{c}{520K}     \\
$L_1$ Loss Weight                 & \multicolumn{2}{c}{1}        \\
FFT Loss Weight                   & \multicolumn{2}{c}{0.1}      \\
Initial Learning Rate             & \multicolumn{2}{c}{1e-3}     \\ 
LR Decay Factor                   & \multicolumn{2}{c}{0.5}      \\
LR Decay Milestones               & \multicolumn{2}{c}{[200K, 300K, 400K, 480K, 500K]} \\
\bottomrule
\end{tabular}
}
\end{table}

\begin{table*}[t]
\caption{Quantitative comparisons of lightweight SR methods trained on DIV2K. The best and second-best results are highlighted in {\color[HTML]{9a0000} red} and {\color[HTML]{3166ff} blue}, respectively.}
\label{tab:VS_on_div2k}
\resizebox{\textwidth}{!}{%
\begin{tabular}{@{}cccccccccccccc@{}}
\toprule
 &
   &
   &
   &
  \multicolumn{2}{c}{Set5} &
  \multicolumn{2}{c}{Set14} &
  \multicolumn{2}{c}{B100} &
  \multicolumn{2}{c}{Urban100} &
  \multicolumn{2}{c}{Manga109} \\ \cmidrule(l){5-14} 
\multirow{-2.5}{*}{Method} &
  \multirow{-2.5}{*}{Source} &
  \multirow{-2.5}{*}{Scale} &
  \multirow{-2.5}{*}{\# Params} &
  PSNR &
  SSIM &
  PSNR &
  SSIM &
  PSNR &
  SSIM &
  PSNR &
  SSIM &
  PSNR &
  SSIM \\ \midrule
  EDSR~\cite{EDSR} &
  CVPRW17 &
   &
  40,730K &
  38.11 &
  0.9602 &
   33.92 &
  0.9195 &
   32.32 &
  0.9013 &
  32.93 &
  0.9351 &
  39.10 &
  0.9773 \\
IMDN~\cite{IMDN} &
  MM19 &
   &
  694K &
  38.00 &
  0.9605 &
  33.63 &
  0.9177 &
  32.19 &
  0.8996 &
  32.17 &
  0.9283 &
  38.88 &
  0.9774 \\
SwinIR~\cite{swinIR} &
  ICCV21 &
   &
  878K &
  38.14 &
  0.9611 &
  33.86 &
  0.9206 &
  32.31 &
  0.9012 &
  32.76 &
  0.9340 &
  39.12 &
  0.9783 \\
ELAN~\cite{ELAN} &
  ECCV22 &
   &
  621K &
  38.17 &
  0.9611 &
  33.94 &
  0.9207 &
  32.30 &
  0.9012 &
  32.76 &
  0.9340 &
  39.11 &
  0.9782 \\
  LKASR~\cite{LKASR} &
  KBS22 &
   &
  947K &
   38.17 &
  0.9611 &
  33.84 &
   0.9204 &
  32.31 &
   0.9014 &
  32.69 &
  0.9339 &
  39.12 &
   0.9779 \\
DiVANet~\cite{DiVANet} &
  PR23 &
   &
  902K &
  38.16 &
   0.9612 &
  33.80 &
  0.9195 &
  32.29 &
  0.9012 &
  32.60 &
  0.9325 &
  39.08 &
  0.9775 \\
SwinIR-NG~\cite{swinNG} &
  CVPR23 &
   &
  1,181K &
  38.17 &
  0.9612 &
  33.94 &
  0.9205 &
  32.31 &
  0.9013 &
  32.78 &
  0.9340 &
  39.20 &
  0.9781 \\
OmniSR~\cite{omni-SR} &
  CVPR23 &
   &
  772K &
  38.22 &
  0.9613 &
  {\color[HTML]{3166ff} 33.98} &
  0.9210 &
  {\color[HTML]{3166ff} 32.36} &
  {\color[HTML]{3166ff} 0.9020} &
  {\color[HTML]{3166ff} 33.05} &
  0.9363 &
  39.28 &
  0.9784 \\
SRFormer~\cite{SRFormer} &
  ICCV23 &
   &
  853K &
  {\color[HTML]{3166ff} 38.23} &
  0.9613 &
  33.94 &
  0.9209 &
  {\color[HTML]{3166ff} 32.36} &
  0.9019 &
  32.91 &
  0.9353 &
  39.28 &
  {\color[HTML]{3166ff} 0.9785} \\
MambaIR~\cite{MambaIR} &
  ECCV24 &
   &
  905K &
  38.13 &
  0.9610 &
  33.95 &
  0.9208 &
  32.31 &
  0.9013 &
  32.85 &
  0.9349 &
  39.20 &
  0.9782 \\
HiT-SIR~\cite{HiT-SR} &
  ECCV24 &
   &
  772K &
  38.22 &
   0.9613 &
  33.91 &
  {\color[HTML]{3166ff} 0.9213} &
  32.35 &
  0.9019 &
  33.02 &
  {\color[HTML]{3166ff} 0.9365} &
   39.38 &
  0.9782 \\
CRAFT~\cite{CRAFT} &
  TPAMI25 &
   &
  738K &
  38.23 &
  {\color[HTML]{9a0000} 0.9615} &
  33.92 &
   0.9211 &
  32.33 &
  0.9016 &
  32.86 &
   0.9343 &
  {\color[HTML]{3166ff} 39.39} &
  {\color[HTML]{9a0000} 0.9786} \\
MaIR~\cite{MaIR} &
  CVPR25 &
   &
  878K &
  38.18 &
  0.9610 &
  33.89 &
  0.9209 &
  32.31 &
  0.9013 &
  32.89 &
  0.9346 &
  39.22 &
  0.9778 \\
EchoSR &
  Ours &
  \multirow{-15}{*}{$\times2$} &
  929K &
  {\color[HTML]{9a0000} 38.26} &
  {\color[HTML]{9a0000} 0.9615} &
  {\color[HTML]{9a0000} 33.99} &
  {\color[HTML]{9a0000} 0.9221} &
  {\color[HTML]{9a0000} 32.38} &
  {\color[HTML]{9a0000} 0.9022} &
  {\color[HTML]{9a0000} 33.09} &
  {\color[HTML]{9a0000} 0.9369} &
  {\color[HTML]{9a0000} 39.42} &
  0.9778 \\ \midrule
  EDSR~\cite{EDSR} &
  CVPRW17 &
   &
  43,680K &
   34.65 &
  0.9280 &
  30.52 &
  0.8462 &
  29.25 &
   0.8093 &
   28.80 &
   0.8653 &
   34.17 &
  0.9476 \\
IMDN~\cite{IMDN} &
  MM19 &
   &
  703K &
  34.36 &
  0.9270 &
  30.32 &
  0.8417 &
  29.09 &
  0.8046 &
  28.17 &
  0.8519 &
  33.61 &
  0.9445 \\
SwinIR~\cite{swinIR} &
  ICCV21 &
   &
  886K &
  34.62 &
  0.9289 &
  30.54 &
  0.8463 &
  29.20 &
  0.8082 &
  28.66 &
  0.8624 &
  33.98 &
  0.9478 \\
ELAN~\cite{ELAN} &
  ECCV22 &
   &
  629K &
  34.61 &
  0.9288 &
  30.55 &
  0.8463 &
  29.21 &
  0.8081 &
  28.69 &
  0.8624 &
  34.00 &
  0.9478 \\
  LKASR~\cite{LKASR} &
  KBS22 &
   &
  947K &
  34.64 &
  0.9278 &
   30.55 &
   0.8464 &
  29.20 &
  0.8077 &
  28.55 &
  0.8604 &
  34.11 &
   0.9477 \\
DiVANet~\cite{DiVANet} &
  PR23 &
   &
  949K &
  34.60 &
   0.9285 &
  30.47 &
  0.8447 &
  29.19 &
  0.8073 &
  28.58 &
  0.8603 &
  33.94 &
  0.9468 \\
SwinIR-NG~\cite{swinNG} &
  CVPR23 &
   &
  1,190K &
  34.64 &
  0.9293 &
  30.58 &
  0.8471 &
  29.24 &
  0.8090 &
  28.75 &
  0.8639 &
  34.22 &
  0.9488 \\
OmniSR~\cite{omni-SR} &
  CVPR23 &
   &
  780K &
  34.70 &
  0.9294 &
  30.57 &
  0.8469 &
  {\color[HTML]{3166ff} 29.28} &
  0.8094 &
  28.84 &
  0.8656 &
  34.22 &
  0.9487 \\
SRFormer~\cite{SRFormer} &
  ICCV23 &
   &
  861K &
  34.67 &
  0.9296 &
  30.57 &
  0.8469 &
  29.26 &
  0.8099 &
  28.81 &
  0.8655 &
  34.19 &
  0.9489 \\
MambaIR~\cite{MambaIR} &
  ECCV24 &
   &
  913K &
  34.63 &
  0.9288 &
  30.54 &
  0.8459 &
  29.23 &
  0.8084 &
  28.70 &
  0.8631 &
  34.12 &
  0.9479 \\
HiT-SIR~\cite{HiT-SR} &
  ECCV24 &
   &
  780K &
  {\color[HTML]{9a0000} 34.72} &
  {\color[HTML]{3166ff} 0.9298} &
  {\color[HTML]{3166ff} 30.62} &
  {\color[HTML]{3166ff} 0.8474} &
  29.27 &
  {\color[HTML]{3166ff} 0.8101} &
  {\color[HTML]{3166ff} 28.93} &
  {\color[HTML]{3166ff} 0.8673} &
  {\color[HTML]{3166ff} 34.40} &
  {\color[HTML]{3166ff} 0.9496} \\
  CRAFT~\cite{CRAFT} &
TPAMI25 &
   &
  744K &
  {\color[HTML]{3166ff}34.71} &
  0.9295 &
  30.61 &
   0.8469 &
  29.24 &
  0.8093 &
  28.77 &
   0.8635 &
   39.29 &
   0.9491 \\
MaIR~\cite{MaIR} &
  CVPR25 &
   &
  886K &
  34.68 &
  0.9292 &
  30.54 &
  0.8461 &
  29.25 &
  0.8088 &
  28.83 &
  0.8651 &
  34.21 &
  0.9484 \\
EchoSR &
  Ours &
  \multirow{-15}{*}{$\times3$} &
  937K &
  {\color[HTML]{3166ff} 34.71} &
  {\color[HTML]{9a0000} 0.9299} &
  {\color[HTML]{9a0000} 30.65} &
  {\color[HTML]{9a0000} 0.8483} &
  {\color[HTML]{9a0000} 29.31} &
  {\color[HTML]{9a0000} 0.8106} &
  {\color[HTML]{9a0000} 28.97} &
  {\color[HTML]{9a0000} 0.8681} &
  {\color[HTML]{9a0000} 34.49} &
  {\color[HTML]{9a0000} 0.9498} \\ \midrule
  EDSR~\cite{EDSR} &
  CVPRW17 &
   &
  43,090K &
   32.46 &
  0.8968 &
  28.80 &
   0.7876 &
   27.71 &
   0.7420 &
   26.64 &
  0.8033 &
  31.02 &
   0.9148 \\
IMDN~\cite{IMDN} &
  MM19 &
   &
  715K &
  32.21 &
  0.8948 &
  28.58 &
  0.7811 &
  27.56 &
  0.7353 &
  26.04 &
  0.7838 &
  30.45 &
  0.9075 \\
SwinIR~\cite{swinIR} &
  ICCV21 &
   &
  897K &
  32.44 &
  0.8976 &
  28.77 &
  0.7858 &
  27.69 &
  0.7406 &
  26.47 &
  0.7980 &
  30.92 &
  0.9151 \\
ELAN~\cite{ELAN} &
  ECCV22 &
   &
  640K &
  32.43 &
  0.8975 &
  28.78 &
  0.7858 &
  27.69 &
  0.7406 &
  26.54 &
  0.7982 &
  30.92 &
  0.9150 \\
LKASR~\cite{LKASR} &
  KBS22 &
   &
  1026K &
   32.46 &
   0.8979 &
   28.84 &
  0.7870 &
  27.71 &
  0.7404 &
  26.54 &
  0.7986 &
  31.01 &
   0.9148 \\
DiVANet~\cite{DiVANet} &
  PR23 &
   &
  939K &
  32.41 &
  0.8973 &
  28.70 &
  0.7844 &
  27.65 &
  0.7391 &
  26.42 &
  0.7958 &
  30.73 &
  0.9119 \\
SwinIR-NG~\cite{swinNG} &
  CVPR23 &
   &
  1,201K &
  32.44 &
  0.8980 &
  28.83 &
  0.7870 &
  {\color[HTML]{3166ff} 27.73} &
  0.7418 &
  26.61 &
  0.8010 &
  31.09 &
  0.9161 \\
OmniSR~\cite{omni-SR} &
  CVPR23 &
   &
  792K &
  32.49 &
  0.8988 &
  28.78 &
  0.7859 &
  27.71 &
  0.7415 &
  26.64 &
  0.8018 &
  31.02 &
  0.9151 \\
SRFormer~\cite{SRFormer} &
  ICCV23 &
   &
  873K &
   32.51 &
  0.8988 &
  28.82 &
  0.7872 &
  {\color[HTML]{3166ff} 27.73} &
  0.7422 &
  26.67 &
  0.8032 &
  31.17 &
  0.9165 \\
MambaIR~\cite{MambaIR} &
  ECCV24 &
   &
  924K &
  32.42 &
  0.8977 &
  28.74 &
  0.7847 &
  27.68 &
  0.7400 &
  26.52 &
  0.7983 &
  30.94 &
  0.9135 \\
HiT-SIR~\cite{HiT-SR} &
  ECCV24 &
   &
  792K &
  32.51 &
  {\color[HTML]{3166ff} 0.8991} &
   28.84 &
  {\color[HTML]{3166ff} 0.7873} &
  {\color[HTML]{3166ff} 27.73} &
  {\color[HTML]{3166ff} 0.7424} &
  {\color[HTML]{3166ff} 26.71} &
  {\color[HTML]{3166ff} 0.8045} &
  {\color[HTML]{3166ff} 31.23} &
  {\color[HTML]{3166ff} 0.9176} \\
CRAFT~\cite{CRAFT} &  
TPAMI25 &
   &
  753K &
  {\color[HTML]{3166ff}32.52} &
  0.8989 &
  {\color[HTML]{3166ff}28.85} &
   0.7872 &
  27.72 &
  0.7418 &
  26.56 &
   0.7995 &
   31.18 &
   0.9168 \\
MaIR~\cite{MaIR} &
  CVPR25 &
   &
  897K &
  32.48 &
  0.8985 &
  28.81 &
  0.7864 &
  27.71 &
  0.7414 &
  26.60 &
  0.8013 &
  31.13 &
  {\color[HTML]{9a0000}
  0.9175} \\
EchoSR &
  Ours &
  \multirow{-15}{*}{$\times4$} &
  948K &
  {\color[HTML]{9a0000} 32.56} &
  {\color[HTML]{9a0000} 0.8995} &
  {\color[HTML]{9a0000} 28.89} &
  {\color[HTML]{9a0000} 0.7882} &
  {\color[HTML]{9a0000} 27.77} &
  {\color[HTML]{9a0000} 0.7429} &
  {\color[HTML]{9a0000} 26.78} &
  {\color[HTML]{9a0000} 0.8057} &
  {\color[HTML]{9a0000} 31.38} &
   0.9171 \\ \bottomrule
\end{tabular}%
}
\end{table*}

\section{Experiments}

\subsection{Experimental Setup}
\textbf{Datasets.} For a comprehensive and fair comparison, following established practices~\cite{MambaIR,swinIR}, we utilized the DIV2K~\cite{DIV2K} and DF2K (which includes Flickr2K~\cite{Flickr2K} and DIV2K) datasets for training, respectively. The DIV2K includes 800 high-resolution images, and the Flicker2K includes 2650 images in the training set. LR images were generated by applying bicubic downsampling to HR images. We assessed our method for performance evaluation on the standard SR benchmark datasets: Set5, Set14, B100, Urban100, and Manga109. Specifically, Set5, Set14, and B100 comprise 5, 14, and 100 images, respectively, featuring portraits and natural scenes. Urban100 contains 100 high-resolution images of urban buildings, while Manga109 consists of 109 cover images from magazines. For real-world super-resolution, we employed the test set of RealSR, which consists of 100 pairs of camera-captured HR-LR images. PSNR and SSIM were computed on the Y channel of the YCbCr color space.

\textbf{Implementation details.}
We randomly extracted crop patches of size $72 \times 72$ to augment the training data. These patches underwent data augmentation, including random horizontal flips and random rotations by multiples of $90\degree$ (i.e., $0\degree$, $90\degree$, $180\degree$, and $270\degree$). Following the previous works~\cite{SeeMore,SAFMN,ShuffleMixer}, we employed the Adam optimizer with an $L_1$ loss in both the pixel and frequency domains, performing 520K iterations with a batch size of 32. The initial learning rate was set to $1 \times 10^{-3}$ and halved at the following milestones: [200K, 300K, 400K, 480K, 500K]. All experiments were conducted on an NVIDIA RTX 4090 GPU using PyTorch. 

We implemented two variants, EchoSR and EchoSR-lite. The EchoSR model comprises 4 CHRGs, each containing five stacked CHBs and a COFB, with a channel width of 60.
EchoSR-lite includes 4 CHRGs, but employs a lighter design by stacking [2, 3, 2, 3] CHBs per group, and with a channel width of 36. 
This lightweight version is trained exclusively on the DF2K dataset.
For both EchoSR and EchoSR-lite, we adopt CAFFN~\cite{MogaNet} as the feed-forward network. The details configuration can be found in Table~\ref{tab:training_details}.

\begin{table*}[t]
\caption{Quantitative comparisons of lightweight SR methods trained on DF2K. The best and second-best results are highlighted in {\color[HTML]{9a0000} red} and {\color[HTML]{3166ff} blue}, respectively.}
\label{tab:VSOnDF2K}
\resizebox{\textwidth}{!}{%
\begin{tabular}{@{}cccccccccccccc@{}}
\toprule
 &
   &
   &
   &
  \multicolumn{2}{c}{Set5} &
  \multicolumn{2}{c}{Set14} &
  \multicolumn{2}{c}{B100} &
  \multicolumn{2}{c}{Urban100} &
  \multicolumn{2}{c}{Manga109} \\ \cmidrule(l){5-14} 
\multirow{-2.5}{*}{Method} &
  \multirow{-2.5}{*}{Source} &
  \multirow{-2.5}{*}{Scale} &
  \multirow{-2.5}{*}{\# Params} &
  PSNR &
  SSIM &
  PSNR &
  SSIM &
  PSNR &
  SSIM &
  PSNR &
  SSIM &
  PSNR &
  SSIM \\ \midrule
DAT~\cite{DAT} &
  ICCV23 &
   &
  553K &
  38.24 &
  0.9614 &
  34.01 &
  {\color[HTML]{3166ff} 0.9214} &
  32.34 &
  0.9019 &
  32.89 &
  0.9346 &
  39.49 &
  {\color[HTML]{3166ff}
  0.9788} \\
OSFFNet~\cite{OSFFNet} &
  AAAI24 &
   &
  516K &
  38.11 &
  0.9610 &
  33.72 &
  0.9190 &
  32.29 &
  0.9012 &
  32.67 &
  0.9331 &
  39.09 &
  0.9780 \\
MAN~\cite{MAN} &
  CVPRW24 &
   &
  820K &
  38.18 &
  0.9612 &
  33.93 &
  0.9213 &
  {\color[HTML]{3166ff} 32.36 } &
  {\color[HTML]{3166ff} 0.9022} &
  {\color[HTML]{3166ff} 32.92} &
  {\color[HTML]{3166ff} 0.9364} &
  39.44 &
  0.9786 \\
SeemoRe~\cite{SeeMore} &
  ICML24 &
   &
  931K &
  {\color[HTML]{3166ff} 38.27} &
  {\color[HTML]{9a0000} 0.9616} &
  {\color[HTML]{3166ff} 34.01} &
  0.9210 &
  32.35 &
  0.9018 &
  32.87 &
  0.9344 &
  {\color[HTML]{9a0000} 39.49} &
  {\color[HTML]{9a0000} 0.9790} \\
SRConvNet~\cite{SRConvNet} &
  IJCV25 &
   &
  885K &
  38.14 &
  0.9610 &
  33.81 &
  0.9199 &
  32.28 &
  0.9010 &
  32.59 &
  0.9321 &
  39.22 &
  0.9779 \\
EchoSR &
  Ours &
  \multirow{-6}{*}{$\times2$} &
  929K &
  {\color[HTML]{9a0000} 38.30} &
  {\color[HTML]{3166ff} 0.9615} &
  {\color[HTML]{9a0000} 34.15} &
  {\color[HTML]{9a0000} 0.9236} &
  {\color[HTML]{9a0000} 32.39} &
  {\color[HTML]{9a0000} 0.9024} &
  {\color[HTML]{9a0000} 33.15} &
  {\color[HTML]{9a0000} 0.9373} &
  {\color[HTML]{9a0000} 39.49} &
   0.9787 \\ \midrule
DAT~\cite{DAT} &
  ICCV23 &
   &
  561K &
  {\color[HTML]{9a0000} 34.76} &
  {\color[HTML]{3166ff} 0.9299} &
  {\color[HTML]{3166ff} 30.63} &
  0.8474 &
  29.29 &
  {\color[HTML]{3166ff} 0.8103} &
  {\color[HTML]{3166ff} 28.89} &
  0.8666 &
  {\color[HTML]{3166ff} 34.55} &
  {\color[HTML]{3166ff} 0.9501} \\
OSFFNet~\cite{OSFFNet} &
  AAAI24 &
   &
  524K &
  34.58 &
  0.9287 &
  30.48 &
  0.8450 &
  29.21 &
  0.8080 &
  28.49 &
  0.8595 &
  34.00 &
  0.9472 \\
MAN~\cite{MAN} &
  CVPRW24 &
   &
  829K &
  34.65 &
  0.9292 &
  30.60 &
  {\color[HTML]{3166ff} 0.8476} &
  29.29 &
  0.8101 &
  28.87 &
  {\color[HTML]{3166ff} 0.8671} &
  34.40 &
  0.9493 \\
SeemoRe~\cite{SeeMore} &
  ICML24 &
   &
  959K &
  34.72 &
  0.9279 &
  30.60 &
  0.8469 &
  {\color[HTML]{3166ff} 29.29} &
  0.8101 &
  28.86 &
  0.8653 &
  34.53 &
  0.9496 \\
SRConvNet~\cite{SRConvNet} &
  IJCV25 &
   &
  906K &
  34.59 &
  0.9288 &
  30.50 &
  0.8455 &
  29.22 &
  0.8081 &
  28.56 &
  0.8600 &
  34.17 &
  0.9479 \\
EchoSR &
  Ours &
  \multirow{-6}{*}{$\times3$} &
  937K &
  {\color[HTML]{3166ff} 34.74} &
  {\color[HTML]{9a0000} 0.9301} &
  {\color[HTML]{9a0000} 30.67} &
  {\color[HTML]{9a0000} 0.8484} &
  {\color[HTML]{9a0000} 29.32} &
  {\color[HTML]{9a0000} 0.8110} &
  {\color[HTML]{9a0000} 29.03} &
  {\color[HTML]{9a0000} 0.8691} &
  {\color[HTML]{9a0000} 34.58} &
  {\color[HTML]{9a0000} 0.9502} \\
  \midrule
DAT~\cite{DAT} &
  ICCV23 &
   &
  573K &
  {\color[HTML]{9a0000} 32.57} &
  {\color[HTML]{3166ff} 0.8991} &
  28.87 &
  0.7879 &
  27.74 &
  0.7428 &
  26.64 &
  0.8033 &
  31.37 &
  {\color[HTML]{3166ff} 0.9178} \\
OSFFNet~\cite{OSFFNet} &
  AAAI24 &
   &
  537K &
  32.39 &
  0.8976 &
  28.75 &
  0.7852 &
  27.66 &
  0.7393 &
  26.36 &
  0.7950 &
  30.84 &
  0.9125 \\
MAN~\cite{MAN} &
  CVPRW24 &
   &
  840K &
  32.50 &
  0.8988 &
  28.87 &
  0.7885 &
  27.77 &
  {\color[HTML]{3166ff} 0.7429} &
  26.70 &
  {\color[HTML]{3166ff} 0.8052} &
  31.25 &
  0.9170 \\
SeemoRe~\cite{SeeMore} &
  ICML24 &
   &
  969K &
  32.51 &
  0.8990 &
  {\color[HTML]{3166ff} 28.92} &
  {\color[HTML]{9a0000} 0.7888} &
  {\color[HTML]{3166ff} 27.78} &
  0.7428 &
  {\color[HTML]{3166ff} 26.79} &
  0.8046 &
  {\color[HTML]{9a0000} 31.48} &
  {\color[HTML]{9a0000} 0.9181} \\
SRConvNet~\cite{SRConvNet} &
  IJCV25 &
   &
  902K &
  32.44 &
  0.8976 &
  28.77 &
  0.7857 &
  27.69 &
  0.7402 &
  26.47 &
  0.7970 &
  30.96 &
  0.9139 \\
EchoSR &
  Ours &
  \multirow{-6}{*}{$\times4$} &
  948K &
  {\color[HTML]{3166ff} 32.55} &
  {\color[HTML]{9a0000} 0.8995} &
  {\color[HTML]{9a0000} 28.93} &
  {\color[HTML]{9a0000} 0.7888} &
  {\color[HTML]{9a0000} 27.80} &
  {\color[HTML]{9a0000} 0.7437} &
  {\color[HTML]{9a0000} 26.83} &
  {\color[HTML]{9a0000} 0.8072} &
  {\color[HTML]{3166ff} 31.43} &
  0.9171 \\ \bottomrule
\end{tabular}%
}
\end{table*}

\subsection{Comparison with State-of-the-Art Methods}

\textbf{Quantitative comparisons.} We present a comprehensive quantitative evaluation for $\times2$, $\times3$, and $\times4$ image super-resolution tasks. To ensure fair and thorough comparisons with SOTA methods, we evaluated the performance of CNN-based, Transformer-based, and Mamba-based methods on the DIV2K and DF2K datasets, respectively.

As shown in Table~\ref{tab:VS_on_div2k}, we present a comparative evaluation of our DIV2K-trained EchoSR model against leading efficient SR methods trained on DIV2K. This includes CNN-based works such as EDSR~\cite{EDSR}, IMDN~\cite{IMDN}, LKASR~\cite{LKASR}, and DiVANet~\cite{DiVANet}, Transformer-based works like SwinIR~\cite{swinIR}, ELAN~\cite{ELAN}, SwinIR-NG~\cite{swinNG}, OmniSR~\cite{omni-SR}, SRFormer~\cite{SRFormer}, HiT-SIR~\cite{HiT-SR}, and CRAFT~\cite{CRAFT} as well as Mamba-based works like MambaIR~\cite{MambaIR} and MaIR~\cite{MaIR}.

\begin{figure*}[!t]
\centering
\includegraphics[width=1\linewidth]{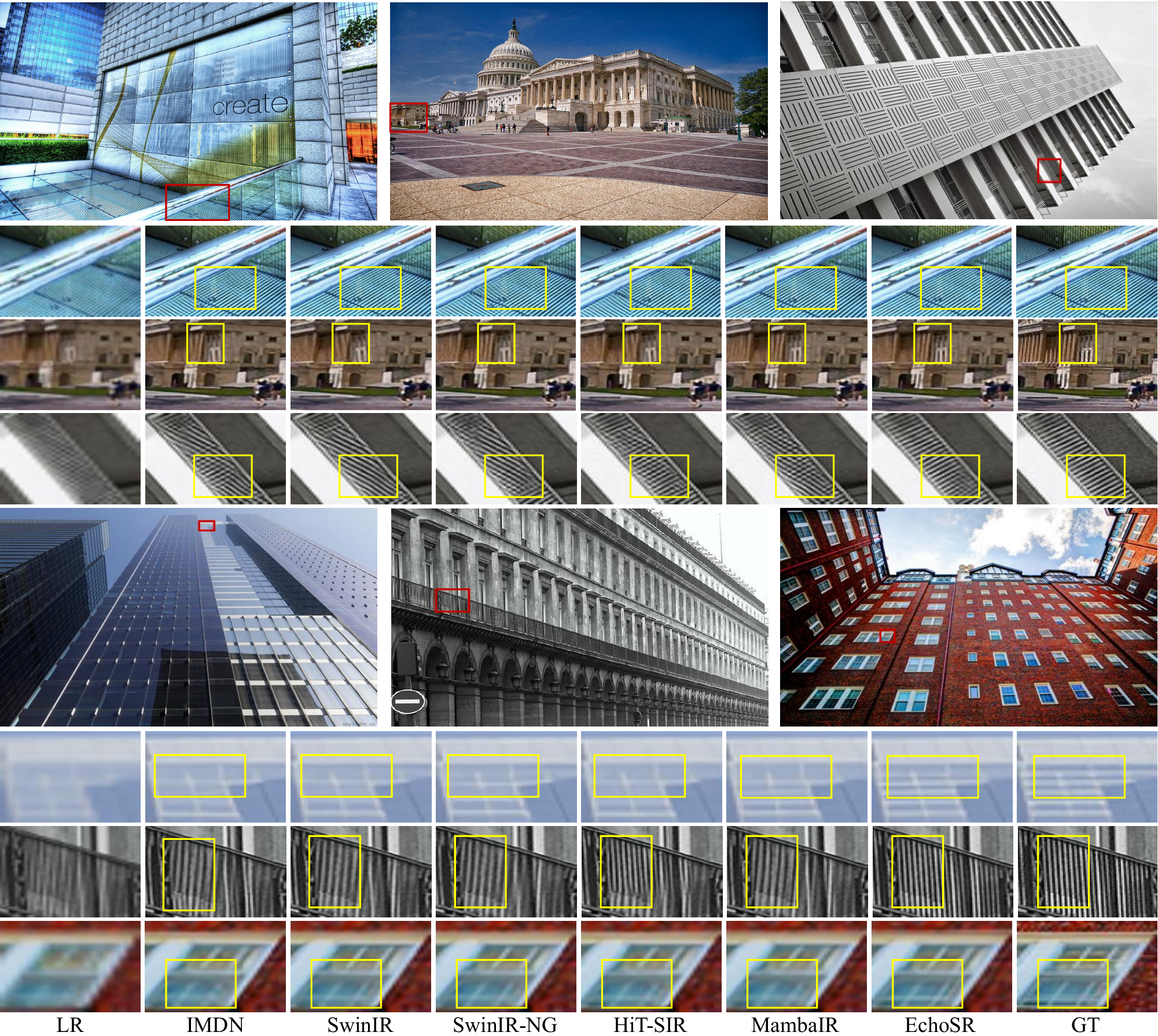}
\caption{Visual comparison of EchoSR (ours) and SOTA methods on the Urban100 benchmark for $\times 2$ SR. EchoSR shows significant advantages in restoring structures and textures, with sharper and clearer results. Please zoom in for a better review. }
\label{fig:urban100x2_2}
\end{figure*}

LKASR stacks dilated convolutions to emulate large-kernel convolutions, it overlooks the importance of multi-scale feature representation. Methods like SRformer and HiT-SIR enhance global features by expanding windows, which, unfortunately, sacrifice efficiency and require more inference memory.
MambaIR and MaIR utilize SSMs for global modeling. However, the auxiliary modules, such as channel attention mechanisms and nested scanning strategies introduced for locality, combined with immature hardware-level optimization, incur additional computational load.
EchoSR surpasses recent SOTA SR approaches by integrating local, multi-scale, and global representations, achieving remarkable PSNR and SSIM scores on nearly all test cases.

In Table~\ref{tab:VSOnDF2K}, all models were trained on the larger-scale DF2K dataset. 
The comparison set includes CNN-based methods OSFFNet~\cite{OSFFNet}, MAN~\cite{MAN}, SeemoRe~\cite{SeeMore}, SRConv~\cite{SRConvNet}, and a Transformer-based approach DAT~\cite{DAT}. 
Table~\ref{tab:VSOnDF2K} illustrates that our EchoSR maintains a leading position when compared against SOTA methods trained on the larger DF2K dataset. Compared with EchoSR, MAN obtains multi-scale large receptive fields through different dilation rates; however, it neglects the local continuity of the image. DAT deeply integrates spatial and channel information and maintains a relatively small number of parameters, but exhibits low actual efficiency. Notably, EchoSR trained on the smaller DIV2K dataset is often capable of delivering near-optimal performance in most scenarios.
Our proposed method consistently exhibits superior performance in most datasets.

\textbf{Qualitative comparisons.} 
Here, we present a visual comparison of our method against several lightweight SR models.
Fig.~\ref{fig:urban100x2_2} visualizes the $\times2$ upscaling SR. The compared SR methods~\cite{IMDN,swinIR,swinNG,HiT-SR,MambaIR} often struggle to generate precise details and suppress artifacts. 
Our EchoSR overcomes these limitations by employing multi-scale receptive field expansion to capture across-scale long-range dependencies, producing reconstructions with enhanced clarity and sharper textures.

\begin{figure*}[!t]
\centering
\includegraphics[width=1\linewidth]{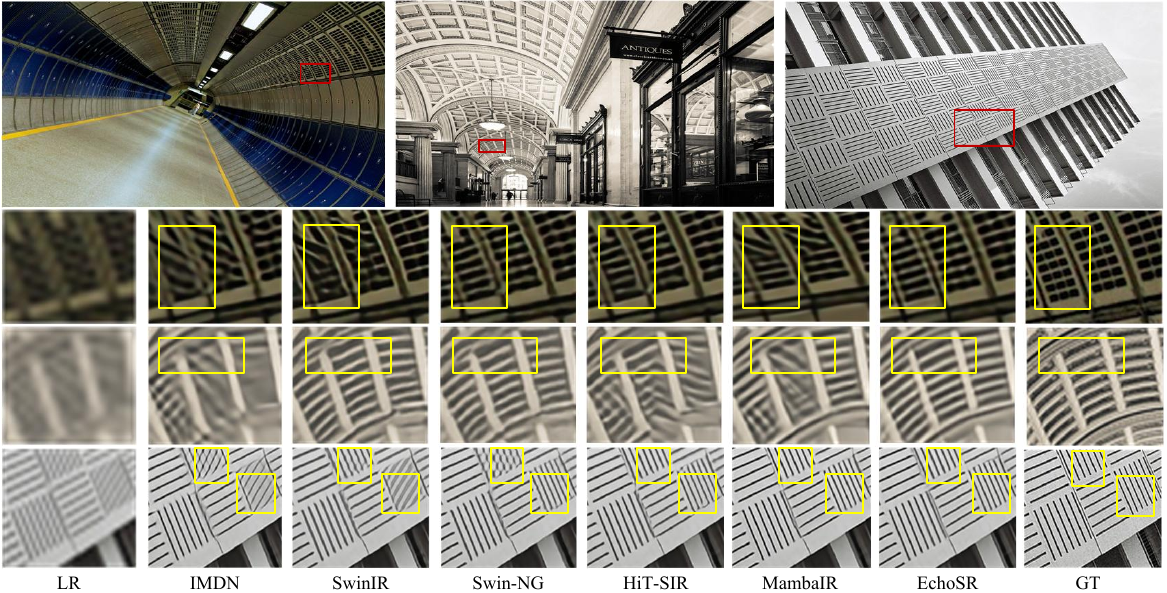}
\caption{Visual comparisons of EchoSR (ours) and SOTA methods on the Urban100 benchmark for $\times 4$ SR. Our method can more effectively restore the structural and texture information. Please zoom in for a better experience.}
\label{fig:urban100x4}
\end{figure*}

\begin{figure*}[!t]
\centering
\includegraphics[width=1\linewidth]{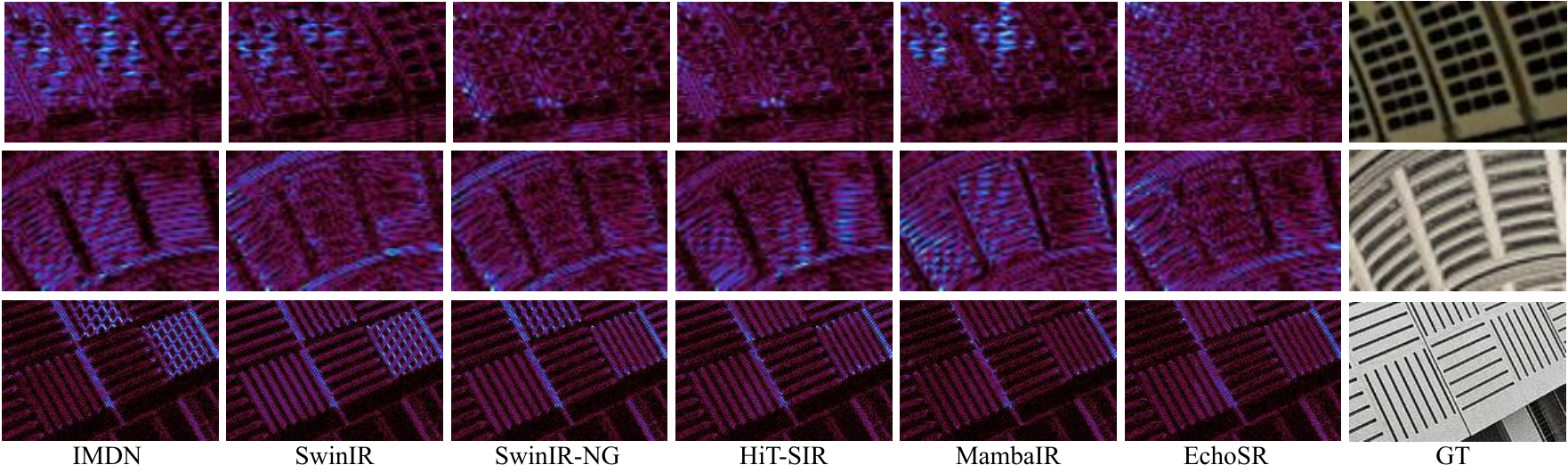}
\caption{Visual comparison of pixel-wise error maps on the Urban100 ($\times4$) dataset. The GT is provided to illustrate the complex high-frequency structures; the consistently minimal residuals (darker regions) produced by EchoSR demonstrate its superior structural fidelity and absence of texture hallucinations.}
\label{fig:urban100x4_error_map}
\end{figure*}

\begin{figure*}[!t]
\centering
\includegraphics[width=1\linewidth]{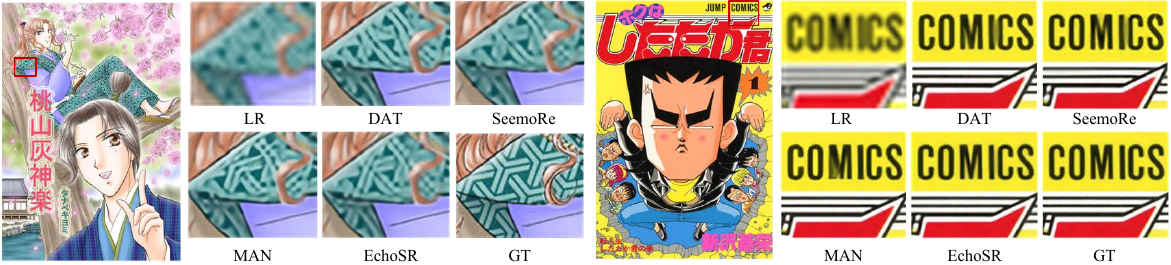}
\caption{Visual comparisons of EchoSR (ours) and SOTA methods on the Manga109 benchmark for $\times 4$ SR. All methods were trained on DF2K. EchoSR reconstructs the textures and structures (\emph{e.g.,} collar and text information) more accurately. }
\label{fig:manga109_x4_DF2K}
\end{figure*}

Fig.~\ref{fig:urban100x4} shows the $\times4$ upscaling SR performance. The accurate preservation of structural integrity, often compromised in SR outputs, remains a significant challenge, especially in complex scenes like the third image in Fig.~\ref{fig:urban100x4}. 
Existing methods~\cite{IMDN,swinIR,swinNG,HiT-SR,MambaIR} frequently fail to faithfully reproduce the underlying image structure, resulting in linear element distortions. 
In contrast, our EchoSR, through a careful balance of global and local contextual cues, achieves both the preservation of overall structural coherence and the precise restoration of intricate linear details. 

This structural fidelity is further corroborated by the pixel-wise error maps provided in Fig.~\ref{fig:urban100x4_error_map}. As observed, EchoSR produces suppressed structural residuals across diverse periodic patterns. Our error maps exhibit no significant artifacts, confirming that the recovered textures are faithful reconstructions. This ensures the reliability and authenticity of the high-resolution outputs.

Fig.~\ref{fig:manga109_x4_DF2K} provides visual results on the Manga109 dataset at $\times 4$ scale, further showcasing the robustness of our approach. 
The effectiveness of EchoSR is further validated through qualitative comparisons with established methods, such as DAT~\cite{DAT}, SeeMoRe~\cite{SeeMore}, and MAN~\cite{MAN}, demonstrating its ability to generate super-resolved images with superior structural fidelity and refined details.

\textbf{Comparison of memory usage and latency.}
Here, we present a comprehensive comparative analysis of memory utilization and inference latency between our method and several SOTA methods, including SwinIR~\cite{swinIR}, SwinIR-NG~\cite{swinNG}, SRformer~\cite{SRFormer}, DAT~\cite{DAT}, HiT-SIR~\cite{HiT-SR}, MambaIR~\cite{MambaIR}, MaIR~\cite{MaIR}, MAN~\cite{MAN}, and SeemoRe~\cite{SeeMore}. To quantify memory usage, we generated random image samples of varying resolutions using PyTorch and subjected them to the models. Inference latency was calculated by averaging the results over 50 iterations.

As shown in Fig.~\ref{psnrvs}, our EchoSR method achieves a PSNR of 33.09 dB on the Urban100 dataset at the $\times2$ scale. This represents a 0.33dB improvement over SwinIR, while also achieving a $\sim10\times$ speedup.
In comparison to SOTA methods such as HiT-SIR and SeemoRe, our approach maintains competitive performance with significantly faster inference speeds.

\begin{figure*}[!t]
\centering
\includegraphics[width=0.495\linewidth]{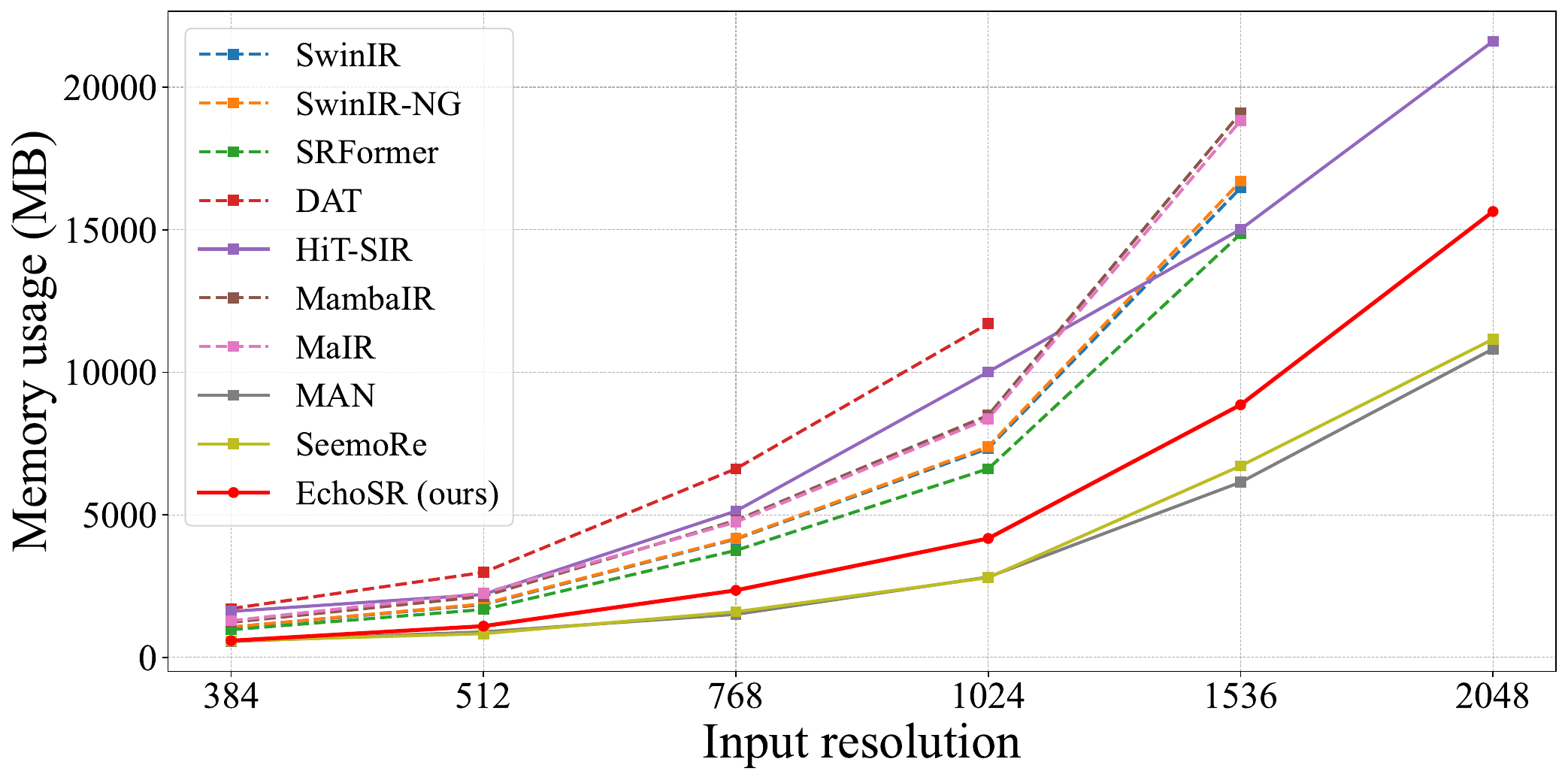}
\includegraphics[width=0.495\linewidth]{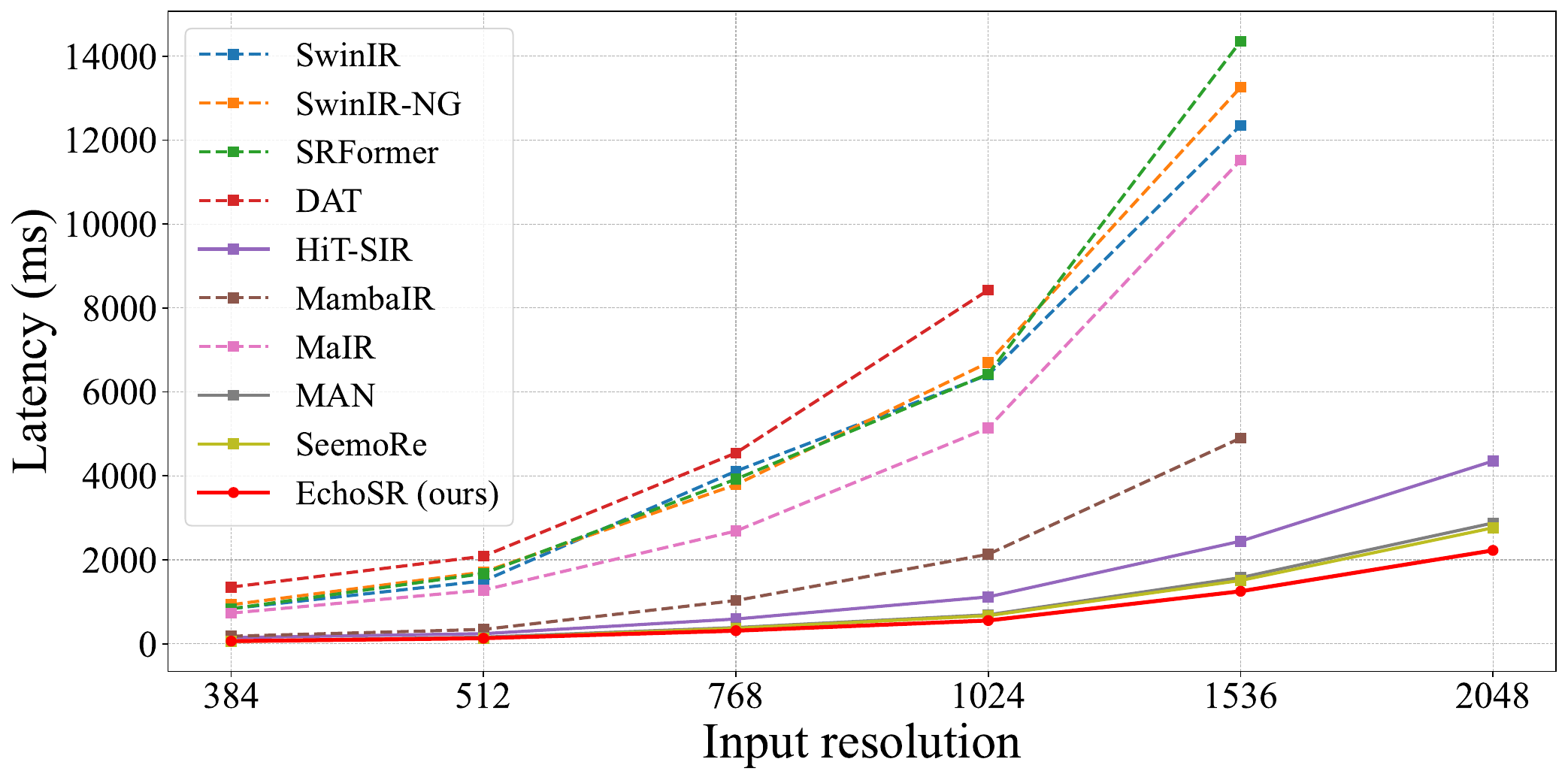}
\caption{Peak GPU memory usage (left) and average inference latency (right) of SR methods at different input resolutions under the $\times 2$ SR setting. EchoSR achieves top performance while maintaining competitive memory efficiency and inference speed.}
\label{fig:compare}
\end{figure*}

Fig.~\ref{fig:compare} (left) shows the peak GPU memory usage on the $\times2$ SR setting. Transformer-based methods such as SwinIR, SwinIR-NG, and SRFormer encountered out-of-memory (OOM) errors when processing inputs of size $3 \times 2048 \times 2048$. Although DAT has the smallest number of parameters among the compared methods, it encounters the OOM problem earlier. The highly optimized HiT-SIR requires more than 21 GB of memory to process $3 \times 2048 \times 2048$ inputs, while our EchoSR only requires about 15 GB, which is slightly higher than the other two CNN-based methods.

\begin{figure}[!t]
\centering
\includegraphics[width=1\linewidth]{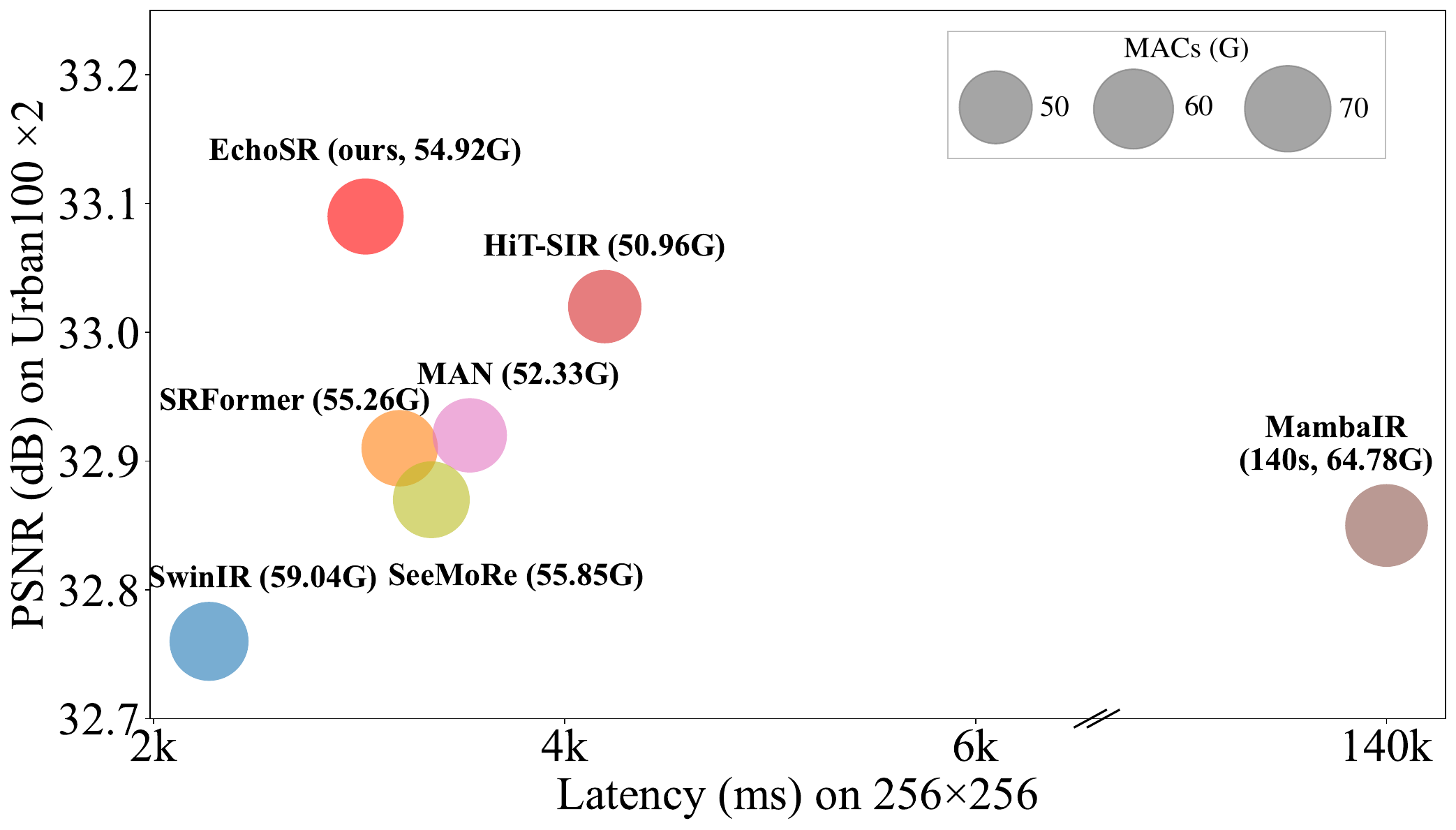}
\caption{Comparisons on the Urban100 test set at $\times2$ scale with input resolution of $256 \times 256$. The area of each circle indicates the computational complexity (MACs). EchoSR (ours) demonstrates a superior trade-off among reconstruction performance, computational efficiency, and inference speed.}
\label{psnrvs_cpu}
\end{figure}

Fig.~\ref{fig:compare} (right) further illustrates the comparison of the average inference latency, demonstrating the efficiency of our method. We note that while MambaIR improves in latency, it is still slower than HiT-SIR; its real-world speed is currently limited by the early stage of SSM kernel optimizations. In addition, MaIR is slower than MambaIR in inference speed due to its nested S-scan mechanism. On average, our EchoSR achieves a remarkable $\sim 10\times$ speedup over the classic SwinIR and a significant $\sim 2\times$ speedup over the SOTA efficient Transformer-based method HiT-SIR. Moreover, EchoSR surpasses other CNN-based methods MAN and SeemoRe by $\sim 1.2\times$.

\begin{figure*}[!t]
\centering
\includegraphics[width=1\linewidth]{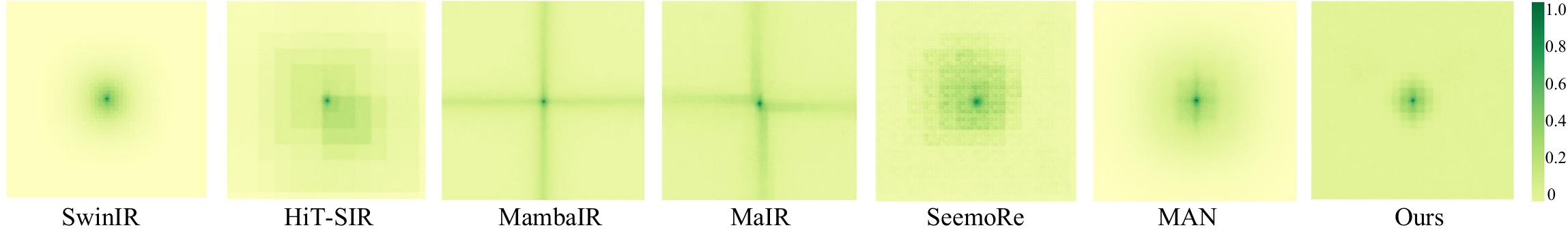}
\caption{Visualization of the ERF across different SR models, including Transformer-based (SwinIR, HiT-SIR), Mamba-based (MambaIR, MaIR), and CNN-based approaches (SeemoRe, MAN, and our EchoSR). Darker regions represent larger ERFs, indicating the extent to which each model captures spatial dependencies. EchoSR demonstrates a broader and more concentrated ERF, reflecting its strong capability in modeling spatial context. }
\label{fig:erf}
\end{figure*}

\textbf{Comparison of latency on mobile SoC.} To evaluate performance in resource-constrained scenarios, we conducted extensive benchmarking on the Snapdragon 8 Elite SoC under a Linux environment. On such edge platforms, inference efficiency is primarily dictated by memory access patterns, including cache locality and memory bandwidth, rather than raw computational throughput. Our results in Fig.~\ref{psnrvs_cpu} reveal a significant discrepancy between mobile CPU latency and the theoretical complexity (MACs) typically emphasized on high-end GPUs. SwinIR achieves superior latency performance due to its streamlined architecture and cache-aligned windowing logic. In contrast, MAN is hindered by the sparse memory access patterns inherent in dilated convolutions, leading to frequent cache misses. Similarly, SeeMore incurs substantial overhead from complex control flow and memory strain caused by its dynamic gating and large-kernel strip convolutions.

For Transformer-based models such as SRFormer and HiT-SIR, frequent high-dimensional tensor permutations and irregular window partitioning create severe data movement bottlenecks in bandwidth-limited mobile environments. Furthermore, MambaIR faces a critical memory challenge, where the recurrent updates of hidden states overwhelm the throughput capacity of mobile caches, a bottleneck exacerbated by the current lack of optimized hardware primitives.

While multi-branch structures often struggle with memory fragmentation, EchoSR mitigates these limitations through a dense parallel convolutional design that avoids the latency overhead of discrete memory access. Experimental results show that EchoSR achieves the second-highest inference speed, surpassed only by SwinIR. Moreover, the peak memory consumption across various resolutions (Fig.~8, left) also underscores our architectural efficiency: the reduced peak memory consumption reflects a more efficient management of intermediate feature maps. This is essential for achieving stable high-throughput inference on edge devices, where excessive memory swapping (DRAM to Cache) often leads to significant latency spikes.

\begin{table}[!t]
\centering
\caption{Comparisons (LPIPS $\downarrow$) with the SOTA methods. The best and second-best results are highlighted in {\color[HTML]{9a0000}red} and {\color[HTML]{3166ff}blue}.}
\label{tab:lpips}
\resizebox{1\columnwidth}{!}{%
\begin{tabular}{@{}ccccccc@{}}
\toprule
Methods &
  Scale &
  Set5 &
  Set14 &
  B100 &
  Urban100 &
  Manga109 \\ \midrule
SwinIR~\cite{swinIR} &
   &
  0.0878 &
  0.1382 &
  0.1573 &
  0.1068 &
  0.0507 \\
SRFormer~\cite{SRFormer} &
   &
  0.0872 &
  0.1372 &
  0.1566 &
  0.1054 &
  {\color[HTML]{3166ff} 0.0499} \\
HiT-SIR~\cite{HiT-SR} &
   &
  {\color[HTML]{2F75B5} 0.0863} &
  {\color[HTML]{3166ff} 0.1369} &
  {\color[HTML]{3166ff} 0.1566} &
  {\color[HTML]{900000} 0.1035} &
  {\color[HTML]{900000} 0.0496} \\
MambaIR~\cite{MambaIR} &
   &
  0.0884 &
  0.1381 &
  0.1584 &
  0.1064 &
  0.0501 \\
MaIR~\cite{MaIR} &
   &
  0.0869 &
  0.1374 &
  0.1579 &
  0.1062 &
  0.0501 \\
EchoSR &
  \multirow{-6}{*}{$\times 2$} &
  {\color[HTML]{900000} 0.0854} &
  {\color[HTML]{900000} 0.1363} &
  {\color[HTML]{900000} 0.1564} &
  {\color[HTML]{3166ff} 0.1038} &
  {\color[HTML]{3166ff} 0.0499} \\ \midrule
SwinIR~\cite{swinIR} &
   &
  0.2071 &
  0.3002 &
  0.3459 &
  0.2786 &
  0.1633 \\
SRFormer~\cite{SRFormer} &
   &
  0.2063 &
  0.3000 &
  0.3439 &
  0.2747 &
  0.1615 \\
HiT-SIR~\cite{HiT-SR} &
   &
  {\color[HTML]{3166ff} 0.2045} &
  0.2974 &
  {\color[HTML]{3166ff} 0.3424} &
  {\color[HTML]{3166ff} 0.2679} &
  {\color[HTML]{3166ff} 0.1582} \\
MambaIR~\cite{MambaIR} &
   &
  0.2074 &
  0.3011 &
  0.3473 &
  0.2771 &
  0.1650 \\
MaIR~\cite{MaIR} &
   &
  0.2046 &
  {\color[HTML]{3166ff} 0.2959} &
  0.3446 &
  0.2706 &
  0.1598 \\
EchoSR &
  \multirow{-6}{*}{$\times 4$} &
  {\color[HTML]{900000} 0.2035} &
  {\color[HTML]{900000} 0.2934} &
  {\color[HTML]{900000} 0.3391} &
  {\color[HTML]{900000} 0.2637} &
  {\color[HTML]{900000} 0.1556} \\ \bottomrule
\end{tabular}%
}
\end{table}

\textbf{Comparison of effective receptive field.} Here, we visualize the effective receptive field (ERF)~\cite{erf} for compared methods. Fig.~\ref{fig:erf} illustrates the ERF visualizations, where darker regions indicate a larger ERF, signifying the extent to which each model captures spatial information. As observed, our EchoSR exhibits a more expansive and uniform ERF compared to methods such as SwinIR~\cite{swinIR} and HiT-SIR~\cite{HiT-SR}, suggesting a superior capacity for capturing global contextual information. Notably, although MambaIR~\cite{MambaIR} adopts a four-directional scanning strategy, its ERF still exhibits a pronounced axis-aligned bias, manifesting as a distinctive cruciform pattern. This suggests that the sequential 1D trajectories, even when oriented in multiple directions, still struggle to achieve the same level of spatial continuity as 2D structures. MaIR~\cite{MaIR}, which utilizes a nested scanning mechanism, shows a similar yet more distorted cruciform distribution. While SeemoRe~\cite{SeeMore} and MAN~\cite{MAN} also exhibit a relatively large ERF, our method displays a more focused and coherent pattern, indicating a more efficient utilization of contextual information.

\textbf{Comparisons of perceptual similarity.} As highlighted in previous studies~\cite{percept}, the often-used PSNR metric does not always align with perceived visual quality, leading to potential discrepancies between quantitative scores and human perception. To evaluate our method's perceptual fidelity, we incorporate the Learned Perceptual Image Patch Similarity (LPIPS) metric. As demonstrated in Table~\ref{tab:lpips}, our EchoSR achieves highly competitive performance, often securing the lowest LPIPS values across various benchmark datasets. These results further substantiate the superiority of our proposed method in terms of perceptual similarity, indicating that EchoSR generates super-resolved images that are not only qualitatively accurate but also visually pleasing and aligned with human perception.

\begin{table*}[!t]
\centering
\caption{Comparisons of tiny SR methods trained on DF2K. The best and second-best results are colored in {\color[HTML]{9a0000}red} and {\color[HTML]{3166ff}blue}. }
\label{tab:tiny}
\resizebox{1.0\textwidth}{!}{%
\begin{tabular}{@{}cccccccccccccc@{}}
\toprule
 &
   &
   &
   &
  \multicolumn{2}{c}{Set5} &
  \multicolumn{2}{c}{Set14} &
  \multicolumn{2}{c}{B100} &
  \multicolumn{2}{c}{Urban100} &
  \multicolumn{2}{c}{Manga109} \\ \cmidrule(l){5-14} 
\multirow{-2.5}{*}{Method} &
  \multirow{-2.5}{*}{Source} &
  \multirow{-2.5}{*}{Scale} &
  \multirow{-2.5}{*}{\# Params} &
  PSNR &
  SSIM &
  PSNR &
  SSIM &
  PSNR &
  SSIM &
  PSNR &
  SSIM &
  PSNR &
  SSIM \\ \midrule
ShuffleMixer~\cite{ShuffleMixer} &
  NIPS22 &
   &
  394K &
  38.01 &
  0.9606 &
  33.63 &
  0.9180 &
  32.17 &
  0.8995 &
  31.89 &
  0.9257 &
  38.83 &
  0.9774 \\
DDistill-SR~\cite{DDistill-SR} &
  TMM23 &
   &
  414K &
  38.03 &
  0.9606 &
  33.61 &
  0.9182 &
  32.19 &
  0.9000 &
  32.18 &
  {\color[HTML]{3166ff} 0.9286} &
  38.94 &
  0.9777 \\
SAFMN~\cite{SAFMN} &
  ICCV23 &
   &
  228K &
  38.00 &
  0.9605 &
  33.54 &
  0.9177 &
  32.16 &
  0.8995 &
  31.84 &
  0.9256 &
  38.71 &
  0.9771 \\
SMFANet~\cite{SMFANet} &
  ECCV24 &
   &
  186K &
  {\color[HTML]{3166ff} 38.08} &
  {\color[HTML]{3166ff} 0.9607} &
  {\color[HTML]{9a0000} 33.65} &
  0.9185 &
  {\color[HTML]{3166ff} 32.22} &
  {\color[HTML]{3166ff} 0.9002} &
  {\color[HTML]{3166ff} 32.20} &
  0.9282 &
  {\color[HTML]{9a0000} 39.11} &
  {\color[HTML]{9a0000} 0.9779} \\
SRConv-T~\cite{SRConvNet} &
  IJCV25 &
   &
  387K &
  38.00 &
  0.9605 &
  33.58 &
  {\color[HTML]{3166ff} 0.9186} &
  32.16 &
  0.8995 &
  32.05 &
  0.9272 &
  38.87 &
  0.9774 \\
EchoSR-lite &
  Ours &
  \multirow{-6}{*}{$\times2$} &
  245K &
  {\color[HTML]{9a0000} 38.10} &
  {\color[HTML]{9a0000} 0.9608} &
  {\color[HTML]{9a0000} 33.65} &
  {\color[HTML]{9a0000} 0.9189} &
  {\color[HTML]{9a0000} 32.23} &
  {\color[HTML]{9a0000} 0.9003} &
  {\color[HTML]{9a0000} 32.22} &
  {\color[HTML]{9a0000} 0.9290} &
  {\color[HTML]{3166ff} 39.07} &
  {\color[HTML]{9a0000} 0.9779} \\ \midrule
ShuffleMixer~\cite{ShuffleMixer} &
  NIPS22 &
   &
  415K &
  34.40 &
  0.9272 &
  30.37 &
  0.8423 &
  29.12 &
  0.8051 &
  28.08 &
  0.8498 &
  33.69 &
  0.9448 \\
DDistill-SR~\cite{DDistill-SR} &
  TMM23 &
   &
  414K &
  34.37 &
  {\color[HTML]{3166ff} 0.9275} &
  30.34 &
  0.8420 &
  29.11 &
  0.8053 &
  28.19 &
  {\color[HTML]{3166ff} 0.8528} &
  33.69 &
  0.9451 \\
SAFMN~\cite{SAFMN} &
  ICCV23 &
   &
  233K &
  34.34 &
  0.9267 &
  30.33 &
  0.8418 &
  29.08 &
  0.8048 &
  27.95 &
  0.8474 &
  33.52 &
  0.9437 \\
SMFANet~\cite{SMFANet} &
  ECCV24 &
   &
  191K &
  {\color[HTML]{3166ff} 34.42} &
  0.9274 &
  {\color[HTML]{3166ff} 30.41} &
  {\color[HTML]{3166ff} 0.8430} &
  {\color[HTML]{9a0000} 29.16} &
  {\color[HTML]{3166ff} 0.8065} &
  {\color[HTML]{9a0000} 28.22} &
  0.8523 &
  {\color[HTML]{9a0000} 33.96} &
  {\color[HTML]{9a0000} 0.9460} \\
SRConv-T~\cite{SRConvNet} &
  IJCV25 &
   &
  387K &
  34.39 &
  0.9272 &
  30.31 &
  0.8413 &
  29.08 &
  0.8047 &
  28.04 &
  0.8498 &
  33.51 &
  0.9442 \\
EchoSR-lite &
  Ours &
  \multirow{-6}{*}{$\times 3$} &
  250K &
  {\color[HTML]{9a0000} 34.48} &
  {\color[HTML]{9a0000} 0.9279} &
  {\color[HTML]{9a0000} 30.45} &
  {\color[HTML]{9a0000} 0.8445} &
  {\color[HTML]{3166ff} 29.15} &
  {\color[HTML]{9a0000} 0.8066} &
  {\color[HTML]{3166ff} 28.21} &
  {\color[HTML]{9a0000} 0.8533} &
  {\color[HTML]{9a0000} 33.96} &
  {\color[HTML]{9a0000} 0.9460} \\ \midrule
ShuffleMixer~\cite{ShuffleMixer} &
  NIPS22 &
   &
  411K &
  32.21 &
  0.8953 &
  28.66 &
  0.7827 &
  27.61 &
  0.7366 &
  26.08 &
  0.7835 &
  30.65 &
  0.9093 \\
DDistill-SR~\cite{DDistill-SR} &
  TMM23 &
   &
  434K &
  32.23 &
  {\color[HTML]{3166ff} 0.8960} &
  28.62 &
  0.7823 &
  27.58 &
  0.7365 &
  {\color[HTML]{9a0000} 26.20} &
  {\color[HTML]{9a0000} 0.7891} &
  30.48 &
  0.9090 \\
SAFMN~\cite{SAFMN} &
  ICCV23 &
   &
  240K &
  32.18 &
  0.8948 &
  28.60 &
  0.7813 &
  27.58 &
  0.7359 &
  25.97 &
  0.7809 &
  30.43 &
  0.9063 \\
SMFANet~\cite{SMFANet} &
  ECCV24 &
   &
  197K &
  {\color[HTML]{3166ff} 32.25} &
  0.8956 &
  {\color[HTML]{9a0000} 28.71} &
  {\color[HTML]{3166ff} 0.7833} &
  {\color[HTML]{9a0000} 27.64} &
  {\color[HTML]{3166ff} 0.7377} &
  {\color[HTML]{3166ff} 26.18} &
  0.7862 &
  {\color[HTML]{9a0000} 30.82} &
  {\color[HTML]{9a0000} 0.9104} \\
SRConv-T~\cite{SRConvNet} &
  IJCV25 &
   &
  382K &
  32.18 &
  0.8951 &
  28.61 &
  0.7818 &
  27.57 &
  0.7359 &
  26.06 &
  0.7845 &
  30.35 &
  0.9075 \\
EchoSR-lite &
  Ours &
  \multirow{-6}{*}{$\times 4$} &
  257K &
  {\color[HTML]{9a0000} 32.29} &
  {\color[HTML]{9a0000} 0.8961} &
  {\color[HTML]{3166ff} 28.69} &
  {\color[HTML]{9a0000} 0.7835} &
  {\color[HTML]{9a0000} 27.64} &
  {\color[HTML]{9a0000} 0.7381} &
  {\color[HTML]{3166ff} 26.18} &
  {\color[HTML]{3166ff} 0.7874} &
  {\color[HTML]{3166ff} 30.80} &
  {\color[HTML]{9a0000} 0.9104} \\ \bottomrule
\end{tabular}%
}
\end{table*}

\begin{figure*}[t]
\centering
\includegraphics[width=1\linewidth]{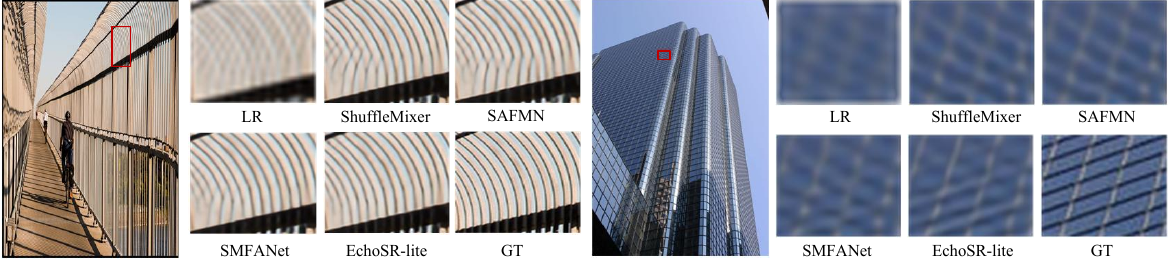}
\caption{Visual comparisons of EchoSR-lite (ours) and SOTA tiny methods on the Urban100 benchmark for $\times 4$ SR. All methods were trained on DF2K. Although $\times4$ SR is a huge challenge for tiny SR methods, our method can still restore faithful details. }
\label{fig:tiny_x4_DF2K}
\end{figure*}

\textbf{Comparison with tiny SR methods.} To further evaluate the efficiency of our approach, we trained a smaller version of our network, EchoSR-lite. We compared it to SOTA, highly optimized tiny SR methods: DDistill-SR\cite{DDistill-SR}, ShuffleMixer\cite{ShuffleMixer}, SAFMN\cite{SAFMN}, SMFANet\cite{SMFANet}, and a reduced parameter version of SRConv\cite{SRConvNet} (SRConv-T). 
All compared methods were trained on DF2K.

As shown in the Table~\ref{tab:tiny}, the compared tiny SR methods often prioritize efficiency through techniques such as distillation~\cite{DDistill-SR} and channel rearrangement~\cite{ShuffleMixer,SAFMN}, which can come at the cost of neglecting the interaction of information flow across different image hierarchies. Our method addresses this limitation by employing pure large-kernel convolutions to achieve an efficient flow of cross-scale information.
EchoSR-lite outperforms the SOTA SMFANet in the tiny parameter regime, demonstrating the effectiveness of our context-harnessing design even with a reduced parameter budget. 

Fig.~\ref{fig:tiny_x4_DF2K} visualizes the $\times4$ upscaling SR results. For tiny SR, it is a challenging task to reconstruct the details. The compared methods~\cite{ShuffleMixer,SAFMN,SMFANet} have difficulty in reconstructing the lines of the first image. In the second image, only our method restores the arrangement direction of the windows.

\begin{table*}[t]
\centering
\caption{Quantitative comparisons of SOTA methods on RealSR (real-world) and Set14, Urban100 (synthetic) datasets. All models were retrained on DF2K using the Real-ESRGAN~\cite{RealESRGAN} degradation pipeline to simulate real-world distortions. The best and second-best results are colored in {\color[HTML]{9a0000}red} and {\color[HTML]{3166ff}blue}. }
\label{tab:realsr}
\resizebox{\textwidth}{!}{%
\begin{tabular}{@{}ccccccccccccc@{}}
\toprule
 &
   &
   &
   &
  \multicolumn{3}{c}{ReaISR} &
  \multicolumn{3}{c}{Set14} &
  \multicolumn{3}{c}{Urban100} \\ \cmidrule(l){5-13} 
\multirow{-2.5}{*}{Methods} &
  \multirow{-2.5}{*}{Source} &
  \multirow{-2.5}{*}{Scale} &
  \multirow{-2.5}{*}{\# Params} &
  PSNR &
  SSIM &
  LPIPS $\downarrow$ &
  PSNR &
  SSIM &
  LPIPS $\downarrow$ &
  PSNR &
  SSIM &
  LPIPS $\downarrow$ \\ \midrule
SwinIR~\cite{swinIR} &
  ICCV21 &
   &
  878K &
  31.99 &
  0.9007 &
  0.2812 &
  28.63 &
  0.8303 &
  0.2798 &
  26.29 &
  0.8248 &
  0.2701 \\
SRFormer~\cite{SRFormer} &
  ICCV23 &
   &
  853K &
  31.90 &
  0.9002 &
  0.2827 &
  28.59 &
  0.8280 &
  0.2811 &
  26.22 &
  0.8208 &
  0.2741 \\
DAT~\cite{DAT} &
  ICCV23 &
   &
  553K &
  32.40 &
  0.9063 &
  0.2737 &
  28.61 &
  0.8315 &
  0.2751 &
  26.13 &
  0.8212 &
  0.2673 \\

  HiT-SR~\cite{HiT-SR} &
  ECCV24 &
   &
  772K &
  32.09 &
  0.9042 &
  0.2738 &
  28.75 &
  0.8373 &
  0.2725 &
  26.27 &
  0.8275 &
  0.2623 \\
MambaIR~\cite{MambaIR} &
  ECCV24 &
   &
  905K &
  32.22 &
  0.9005 &
  0.2751 &
  28.41 &
  0.8202 &
  0.2826 &
  25.74 &
  0.8028 &
  0.2814 \\
SeeMoRe~\cite{SeeMore} &
  ICML24 &
   &
  931K &
  {\color[HTML]{3166ff} 32.41} &
  {\color[HTML]{3166ff} 0.9074} &
  {\color[HTML]{3166ff} 0.2668} &
  {\color[HTML]{900000} 29.08} &
  {\color[HTML]{3166ff} 0.8481} &
  {\color[HTML]{3166ff} 0.2611} &
  {\color[HTML]{900000} 26.66} &
  {\color[HTML]{3166ff} 0.8370} &
  {\color[HTML]{3166ff} 0.2558} \\
EchoSR &
  Ours &
  \multirow{-7}{*}{$\times 2$} &
  929K &
  {\color[HTML]{900000} 32.60} &
  {\color[HTML]{900000} 0.9121} &
  {\color[HTML]{900000} 0.2527} &
  {\color[HTML]{3166ff} 29.06} &
  {\color[HTML]{900000} 0.8551} &
  {\color[HTML]{900000} 0.2490} &
  {\color[HTML]{3166ff} 26.51} &
  {\color[HTML]{900000} 0.8375} &
  {\color[HTML]{900000} 0.2463} \\ \midrule
SwinIR~\cite{swinIR} &
  ICCV21 &
   &
  897K &
  28.06 &
  0.8028 &
  0.3878 &
  26.33 &
  0.7118 &
  0.3905 &
  23.71 &
  0.6940 &
  0.3932 \\
SRFormer~\cite{SRFormer} &
  ICCV23 &
   &
  873K &
  28.07 &
  0.805 &
  0.3869 &
  26.33 &
  0.7083 &
  0.3962 &
  23.68 &
  0.6915 &
  0.3985 \\
DAT~\cite{DAT} &
  ICCV23 &
   &
  573K &
  {\color[HTML]{3166ff} 28.29} &
  0.8064 &
  0.3812 &
  26.33 &
  0.7147 &
  0.3840 &
  23.62 &
  0.6904 &
  0.3886 \\
  HiT-SR~\cite{HiT-SR} &
  ECCV24 &
   &
  792K &
  28.13 &
  {\color[HTML]{3166ff} 0.8074} &
  0.3799 &
  26.45 &
  {\color[HTML]{3166ff} 0.7222} &
  0.3794 &
  23.76 &
  0.7000 &
  0.3821 \\
MambaIR~\cite{MambaIR} &
  ECCV24 &
   &
  924K &
  28.18 &
  0.8017 &
  0.3824 &
  26.18 &
  0.7086 &
  0.3884 &
  23.33 &
  0.6755 &
  0.3965 \\
SeeMoRe~\cite{SeeMore} &
  ICML24 &
   &
  969K &
  {\color[HTML]{900000} 28.37} &
  0.8070 &
  {\color[HTML]{3166ff} 0.3794} &
  {\color[HTML]{3166ff} 26.46} &
  0.7214 &
  {\color[HTML]{3166ff} 0.3756} &
  {\color[HTML]{3166ff} 23.84} &
  {\color[HTML]{3166ff} 0.7015} &
  {\color[HTML]{3166ff} 0.3794} \\
EchoSR &
  Ours &
  \multirow{-7}{*}{$\times 4$} &
  948K &
  {\color[HTML]{3166ff} 28.29} &
  {\color[HTML]{900000} 0.8103} &
  {\color[HTML]{900000} 0.3708} &
  {\color[HTML]{900000} 26.86} &
  {\color[HTML]{900000} 0.7345} &
  {\color[HTML]{900000} 0.3650} &
  {\color[HTML]{900000} 24.16} &
  {\color[HTML]{900000} 0.7139} &
  {\color[HTML]{900000} 0.3663} \\ \bottomrule
\end{tabular}%
}
\end{table*}

\begin{figure*}[!t]
\centering
\includegraphics[width=1\linewidth]{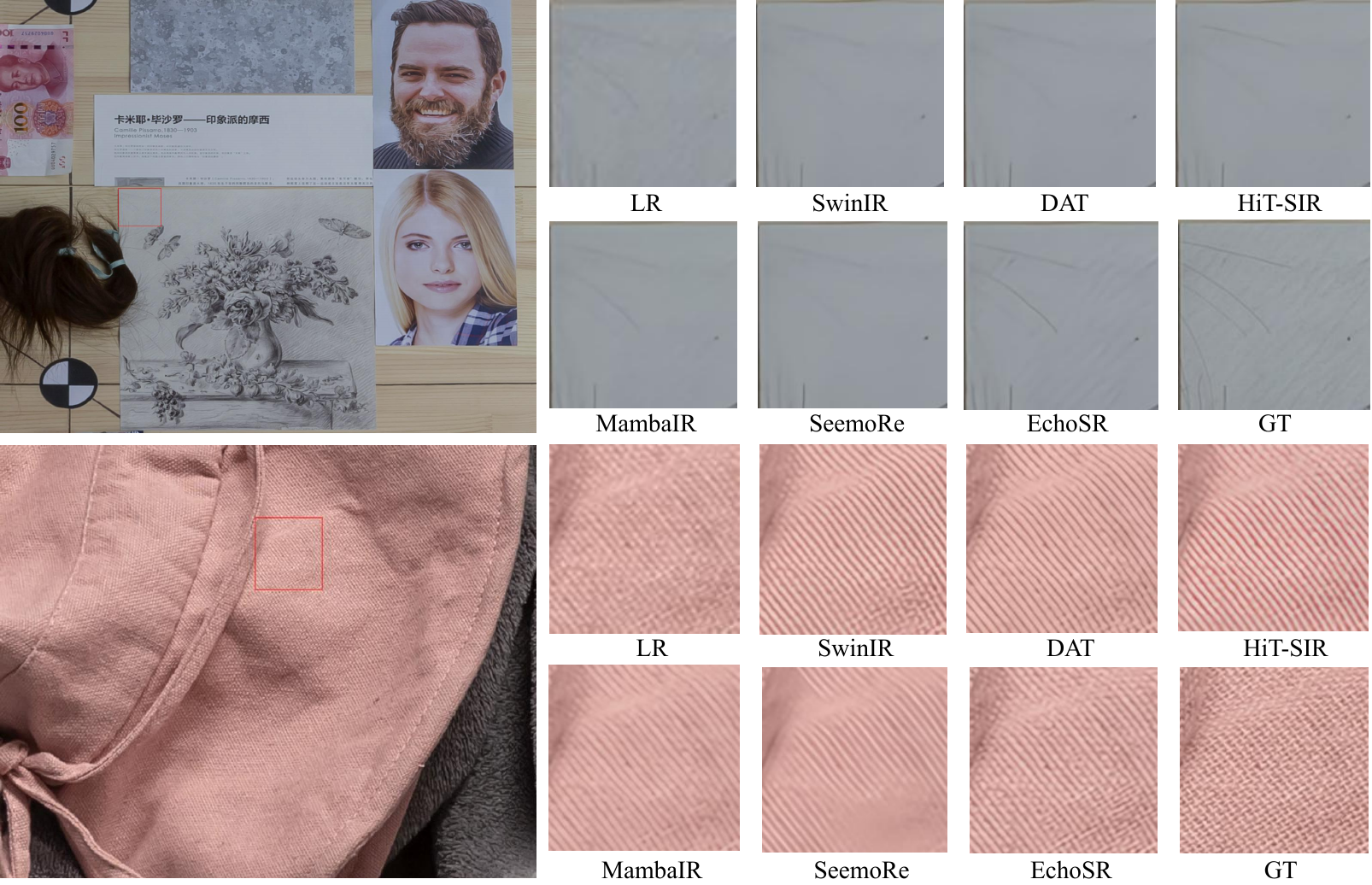}
\caption{Visual comparisons of EchoSR (ours) and SOTA methods on the RealSR dataset for $\times 2$ SR. All methods were retrained on DF2K. Most competing methods tend to generate overly smooth results with texture loss, whereas our method preserves faithful details. }
\label{fig:real_x2_DF2K}
\end{figure*}

\textbf{Real-world super-resolution.} Addressing real-world super-resolution remains a formidable challenge. Predominant SR methods typically rely on fixed degradation kernels (e.g., bicubic downsampling) for training and testing, limiting their generalization capability against unpredictable real-world degradations. To establish a fair comparison benchmark, we retrained EchoSR and other SOTA methods, specifically SwinIR~\cite{swinIR}, SRFormer~\cite{SRFormer}, DAT~\cite{DAT}, HiT-SIR~\cite{HiT-SR}, MambaIR~\cite{MambaIR}, and SeeMoRe~\cite{SeeMore} on the DF2K dataset. We adopted the Real-ESRGAN~\cite{RealESRGAN} training pipeline, which synthesizes realistic degradations via random combinations of blur, resizing, noise, and compression. Evaluations were conducted on the real-world dataset RealSR, as well as synthetic datasets Set14 and Urban100. To account for human visual perception more accurately, we incorporated LPIPS alongside conventional PSNR and SSIM metrics.

Quantitative results in Table~\ref{tab:realsr} demonstrate that EchoSR secures the leading position in most test cases, surpassing the runner-up method by a significant margin in LPIPS. 
This underscores the effectiveness of our local-global-multi-scale architecture in capturing multi-level degradation cues within complex scenes. 
For example, our method achieves an SSIM of 0.9121, surpassing the second-best method, SeeMoRe (0.9074) on the $\times 2$ setting.
This confirms the robust ability of our method to strike a balance between handling simple synthetic patterns and complex real-world distortions.
Visual comparisons on RealSR are presented in Fig.~\ref{fig:real_x2_DF2K}. While competing methods often suffer from over-smoothing when processing intricate textures and degradations, our method excels at reconstructing faithful and sharp details.

\begin{table*}[t]
\centering
\caption{Ablation experiments on different variants of LA: (\#1) removes the  LA module; (\#2) uses 6 CHBs, based on (\#1) configuration; (\#3) removes group convolutions; (\#4) replaces group convolutions with depthwise convolutions; (\#5) removes channel expansion; (\#6) is our full model;
The best results are highlighted in {\color[HTML]{9a0000} red}.}
\label{tab:Ab-LA}
\resizebox{\textwidth}{!}{%
\begin{tabular}{@{}clccccccccccc@{}}
\toprule
\multicolumn{1}{c}{} &
   &
   &
  \multicolumn{2}{c}{Set5} &
  \multicolumn{2}{c}{Set14} &
  \multicolumn{2}{c}{B100} &
  \multicolumn{2}{c}{Urban100} &
  \multicolumn{2}{c}{Manga109} \\ \cmidrule(l){4-13} 
{\multirow{-2.5}{*}{\#}} &
{\multirow{-2.5}{*}{Variant}} &
  \multirow{-2.5}{*}{\# Params} &
  PSNR &
  SSIM &
  PSNR &
  SSIM &
  PSNR &
  SSIM &
  PSNR &
  SSIM &
  PSNR &
  SSIM \\ \midrule

1& w/o LA &
  617K &
  38.14 &
  0.9610 &
  33.82 &
  0.9207 &
  32.28 &
  0.9011 &
  32.64 &
  0.9328 &
  38.97 &
  0.9769 \\
2& 6 CHBs w/o LA &
  878K &
  38.16 &
  0.9612 &
  33.90 &
  0.9612 &
  32.32 &
  0.9015 &
  32.81 &
  0.9351 &
  39.27 &
  0.9780 \\
3& w/o Conv &
  835K &
  38.21 &
  0.9613 &
  33.91 &
  0.9208 &
  32.35 &
  0.9019 &
  32.91 &
  0.9353 &
  39.28 &
  {\color[HTML]{9a0000} 0.9783} \\
4& With DWConv &
  853K &
  38.24 &
  0.9613 &
  {\color[HTML]{9a0000} 34.04} &
  0.9219 &
  32.33 &
  0.9018 &
  32.97 &
  0.9363 &
  39.36 &
  0.9776 \\
5& Fixed channels &
  830K &
  38.22 &
  0.9613 &
  33.88 &
  0.9202 &
  32.34 &
  0.9019 &
  32.91 &
  0.9357 &
  39.31 &
  0.9779 \\ 
  6 &Ours &
  929K &
  {\color[HTML]{9a0000} 38.26} &
  {\color[HTML]{9a0000}0.9615} &
  33.99 &
  {\color[HTML]{9a0000} 0.9221} &
  {\color[HTML]{9a0000} 32.38} &
  {\color[HTML]{9a0000} 0.9022} &
  {\color[HTML]{9a0000} 33.09} &
  {\color[HTML]{9a0000} 0.9369} &
  {\color[HTML]{9a0000} 39.42} &
  0.9778 \\
  \bottomrule
\end{tabular}%
}
\end{table*}

\subsection{Ablation Study}

We conducted a comprehensive ablation study on individual components within our proposed method. All experiments were performed on the $\times2$ super-resolution task using the DIV2K dataset for training. 

\begin{table*}[t]
\centering
\caption{Ablation experiments on GCP and variants of MRFE: (\#1) removes the GCP module; (\#2) replaces the MRFE with a $3 \times 3$ Conv; (\#3) replaces the MRFE with a $11 \times 11$ DWConv; (\#4) removes the identity mapping branch in MRFE module; (\#5) is our full model. 
The $k$ means kernel size. The best results are highlighted in {\color[HTML]{9a0000} red}. }
\label{tab:Ab-MRFE}
\resizebox{\textwidth}{!}{%
\begin{tabular}{@{}llccccccccccc@{}}
\toprule
\multicolumn{1}{c}{} &
   &
   &
  \multicolumn{2}{c}{Set5} &
  \multicolumn{2}{c}{Set14} &
  \multicolumn{2}{c}{B100} &
  \multicolumn{2}{c}{Urban100} &
  \multicolumn{2}{c}{Manga109} \\ \cmidrule(l){4-13} 
{\multirow{-2.5}{*}{\#}} &
{\multirow{-2.5}{*}{Variant}} &
  \multirow{-2.5}{*}{\# Params} &
  PSNR &
  SSIM &
  PSNR &
  SSIM &
  PSNR &
  SSIM &
  PSNR &
  SSIM &
  PSNR &
  SSIM \\ \midrule
1& w/o GCP &
  847K &
  38.23 &
  0.9614 &
  33.96 &
  0.9210 &
  32.34 &
  0.9018 &
  33.00 &
  0.9364 &
  39.35 &
  0.9783 \\  
2& Conv, $k=3$ &
  1,452K &
  {\color[HTML]{9a0000}38.27} &
  0.9614 &
  {\color[HTML]{9a0000}34.03} &
  0.9210 &
  32.36 &
  0.9020 &
  33.05 &
  0.9362 &
  39.37 &
  {\color[HTML]{9a0000} 0.9785} \\
3& DWConv, $k=11$ &
  949K &
  38.23 &
  0.9613 &
  33.86 &
  0.9204 &
  32.35 &
  0.9020 &
  32.99 &
  0.9362 &
  39.28 &
  0.9780 \\
4& w/o identity &
  978K &
  38.24 &
  0.9613 &
  33.93 &
  0.9212 &
  32.35 &
  0.9021 &
  33.01 &
  0.9360 &
  39.32 &
  0.9778 \\
5 &Ours &
  929K &
  38.26 &
  {\color[HTML]{9a0000} 0.9615} &
    33.99 &
  {\color[HTML]{9a0000} 0.9221} &
  {\color[HTML]{9a0000} 32.38} &
  {\color[HTML]{9a0000} 0.9022} &
  {\color[HTML]{9a0000} 33.09} &
  {\color[HTML]{9a0000} 0.9369} &
  {\color[HTML]{9a0000} 39.42} &
  0.9778 \\
  \bottomrule
\end{tabular}%
}
\end{table*}

\begin{figure*}[t]
\centering
\includegraphics[width=1\linewidth]{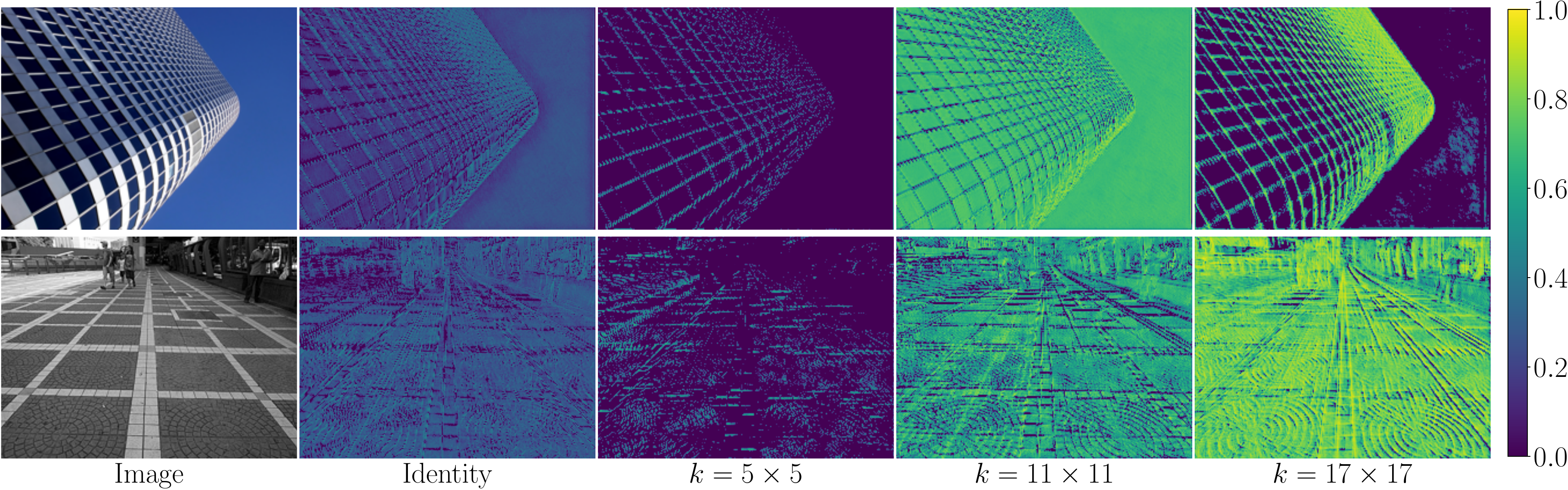}
\caption{Visualization of feature maps in our MRFE module. We showcase outputs from different branches within MRFE. The identity mapping branch retains original features; the $5 \times 5$ kernel branch emphasizes local texture details, while the $11 \times 11$ and $17 \times 17$ kernel branches focus more on capturing global contextual structures. This demonstrates the effectiveness of MRFE in modeling both fine-grained and large-scale spatial information.}
\label{fig:act_map}
\end{figure*}

\textbf{Effectiveness of the LA.} 
As shown in Table~\ref{tab:Ab-LA}, we conduct ablation experiments on different variants of LA. 
We first evaluated the impact of removing the LA (\#1). 
The ablation of the LA resulted in a significant decline across all performance metrics. Subsequently, we increased the number of CHBs to 6 in each group to maintain a comparable parameter count (\#2); however, the PSNR still decreased by 0.28dB and 0.15dB on the Urban100 and Manga109 datasets, respectively. 
In configuration (\#3), we removed the group convolutions, retaining only two linear layers. Configuration (\#4) employed depthwise convolutions; while a marginal performance gain was observed on Set14, the overall trend indicated a decline, particularly pronounced on the highly complex Urban100 dataset. Configuration (\#5), devoid of channel expansion, exhibited performance comparable to (\#3), which lacked group convolutions.
Our LA module effectively extracts local features in high dimensions through channel expansion and the utilization of group convolutions to achieve the best performance, as shown in configuration (\#6).

\textbf{Effectiveness of the GCP.} 
In Table~\ref{tab:Ab-MRFE} (\#1 and \#5), we show the effects of using and without using the GCP module in our full model.
We first tested the importance of the GCP module (\#1), which complements the multi-scale features extracted by MRFE. 
Compared with our full model (\#5), removing GCP led to a PSNR drop of 0.09 dB on Urban100 and 0.07 dB on Manga109. Since these datasets rely more on global structure, the result shows that GCP plays a key role in capturing long-scale structural prior and boosting overall performance.

\textbf{Effectiveness of the MRFE.} 
Table~\ref{tab:Ab-MRFE} (from \#2 to \#5) shows different convolution strategies within our MRFE module.
Configuration (\#2) uses standard convolution with a $3 \times 3$ kernel, reaching 38.27 dB on Set5 and 34.03 dB on Set14, but with a high parameter count of 1,452K. In contrast, configuration (\#3) uses depthwise convolution with an $11 \times 11$ kernel, reducing parameters to 949K (similar to our final EchoSR model) but significantly hurting performance.
In addition, removing the identity mapping branch (\#4), changing the structure from four branches to three, caused a clear performance drop, showing the value of preserving the original feature path. 

\begin{table}[h]
\centering
\caption{Ablation study of kernel configurations and multi-scale design in MRFE (evaluated using PSNR in dB). The 'id' means identity mapping. (\#1-\#6) replace kernel configurations of MRFE;
(\#7) replaces each convolution with strip convolution by decomposing large kernels ($11\times11 \rightarrow 1\times11$ and $11\times1$; $17\times17 \rightarrow 1\times17$ and $17\times1$); (\#8) replaces dilated convolutions with a dilation rate of 3; (\#9) is our full model, achieving the best performance.
The best results are highlighted in {\color[HTML]{9a0000}red}. }
\label{tab:more_kernel}
\resizebox{1\columnwidth}{!}{%
\begin{tabular}{@{}llcccccc@{}}
\toprule
\# & Variant              & \# Params & Set14 & B100  & Urban100                     & Manga109  & Average \\ \midrule
1  & $k = \text{id}, 3, 3, 3 $           & 812K   & 33.84 & 32.34 & 32.94                        & 39.30 & 34.60\\
2  & $k = \text{id}, 5, 5, 5$           & 826K   & 33.90 & 32.36 & 33.01                        & 39.33 & 34.65\\
3  & $k = \text{id}, 11, 11, 11  $      & 913K   & 33.93 & 32.37 & {\color[HTML]{900000} 33.11} & 39.35 & 34.69 \\
4  & $k = \text{id}, 17, 17, 17 $       & 1,064K  & 33.95 & 32.35 & 32.99                        & 39.28 & 34.64\\
5  & $k = \text{id}, 5, 7, 9 $       & 850K  & 33.90 & 32.35 & 33.00                        & 39.29 & 34.63 \\
6  & $k = \text{id}, 5, 15, 25 $       & 1,067K  & 33.96 & 32.36 & 33.03                        & 39.38 & 34.68\\
7  & Strip Conv & 821K   & 33.98 & 32.35 & 32.96                        & 39.28  & 34.64\\
8  & Dilated Conv & 929K   &  33.98 & 32.36 & 32.99                        & 39.36 & 34.67\\
9 & Ours & 929K & {\color[HTML]{900000}33.99} & {\color[HTML]{900000} 32.38} & 33.09 & {\color[HTML]{900000} 39.42}  & {\color[HTML]{900000}34.72}\\ \bottomrule
\end{tabular}%
}
\end{table}

Fig.~\ref{fig:act_map} further illustrates activation maps of each branch from our full model (\#5). It shows that each branch focuses on different fine-grained patterns, highlighting the effectiveness of the MRFE in capturing diverse contextual features.
Our full model, which uses kernel sizes [id, 5, 11, 17], delivers the best balance between efficiency and performance across all datasets, as demonstrated in Table~\ref{tab:Ab-MRFE} (\#5).
  
\textbf{Impacts of kernel configurations.}
Here, we further investigate the impact of using different kernel sizes and multi-scale strategies within the multi-branch structure on the MRFE. 
As shown in Table~\ref{tab:more_kernel}, in configuration (\#1), each branch uses small kernels of size 3. 
Configurations (\#2), (\#3), and (\#4) gradually increase the kernel sizes, and robust performance is achieved when the kernel size is 11 (\#3), followed by a significant drop when the kernel size is increased to 17 (\#4).  
Configurations (\#5) and (\#6) further explore mixed kernel sizes with intervals of 2 and 10, respectively. Configuration (\#5), with its closely spaced kernels, may suffer from a limited receptive field. In contrast, while Configuration (\#6) expands the receptive field with a wider interval, the excessive span between scales may lead to the loss of intermediate feature representations.
Configuration (\#7) in Table~\ref{tab:more_kernel} utilizes strip convolutions. 
Similar to~\cite{InceptionNeXt}, we divided the MRFE into 6 branches, where the large kernels (11 and 17) are decomposed into  $1\times k,k\times1$ forms. Although~\cite{InceptionNeXt,SLaK} have demonstrated the effectiveness of strip convolutions in high-level vision tasks, their decomposition into one-dimensional kernels may compromise the effective modeling of intricate 2D structures and fine-grained textures necessary for high-quality super-resolution. 
Meanwhile, (\#8) used a dilated convolution with a dilation rate of 3, which performed worse than ours (\#9). 
Our full model (\#9), which employs a multi-scale receptive field using pure depthwise convolutions, demonstrates the impressive performance across multiple benchmark datasets. 

\begin{figure}[t]
\centering
\includegraphics[width=1\linewidth]{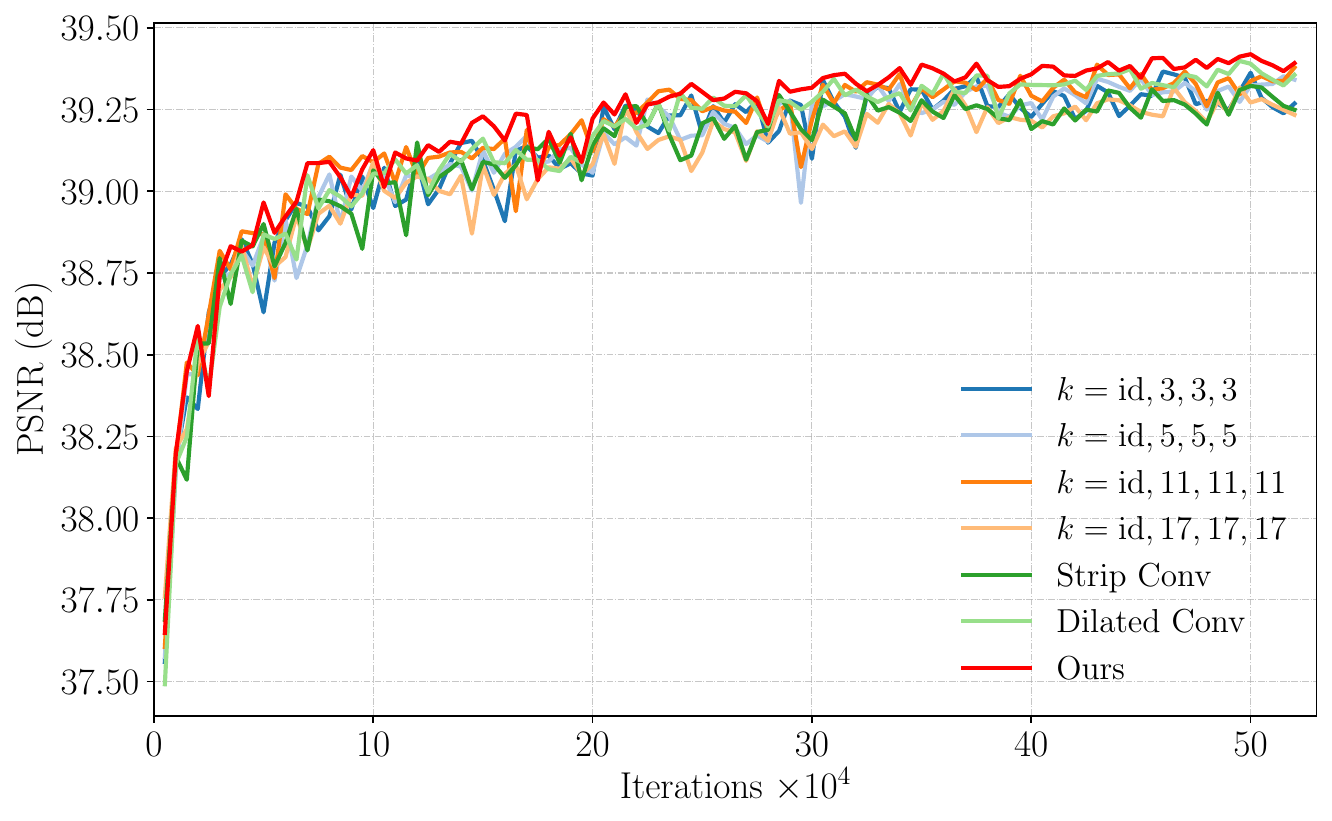}
\caption{Ablation study of kernel configurations and multi-scale design in MRFE. We show training evaluation curves on the Manga109 dataset with different kernel configurations. Our design strategy remains optimal during training.}
\label{fig:m109_curve}
\end{figure}

Fig.~\ref{fig:m109_curve} shows the training evaluation curves on the Manga109 dataset for the ablation study with different kernel configurations. The multi-scale kernel size configuration [id, 5, 11, 17] employed in our method consistently yields faster convergence and superior performance. This outcome not only indicates the improved training stability of our approach but also underscores the effectiveness of large-kernel convolutions in capturing multi-scale features.

\begin{table*}[h]
\centering
\caption{Ablation study of the COFB module. (\#1) replaces the COFB with a standard $3 \times 3$ convolution. (\#2, \#3) uses a single depthwise convolution with kernel size of $7 \times 7$ and $15 \times 15$, respectively. (\#4),(\#5) and (\#6) apply alternative kernel configurations to replace COFB. (\#7) adopts a depthwise dilated convolution with a dilation rate of 3. (\#8) corresponds to our full model using the proposed COFB.
Here, $k$ denotes the kernel size.
The best results are highlighted in {\color[HTML]{9a0000}red}.
}
\label{tab:Ab-COFB}
\resizebox{\textwidth}{!}{%
\begin{tabular}{@{}llccccccccccc@{}}
\toprule
\multicolumn{1}{c}{} &
   &
   &
  \multicolumn{2}{c}{Set5} &
  \multicolumn{2}{c}{Set14} &
  \multicolumn{2}{c}{B100} &
  \multicolumn{2}{c}{Urban100} &
  \multicolumn{2}{c}{Manga109} \\ \cmidrule(l){4-13} 
{\multirow{-2.5}{*}{\#}} &
{\multirow{-2.5}{*}{Variant}} &
  \multirow{-2.5}{*}{\# Params} &
  PSNR &
  SSIM &
  PSNR &
  SSIM &
  PSNR &
  SSIM &
  PSNR &
  SSIM &
  PSNR &
  SSIM \\ \midrule
1& Conv, $k=3$ &
  954K &
  38.22 &
  0.9613 &
  32.90 &
  0.9206 &
  32.35 &
  0.9019 &
  32.97 &
  0.9357 &
  39.26 &
  0.9771 \\
2& DWConv, $k=7$ &
  880K &
  38.24 &
  0.9614 &
  33.92 &
  0.9219 &
  32.36 &
  0.9021 &
  33.01 &
  0.9364 &
  39.31 &
  0.9770 \\
3& DWConv, $k=15$ &
  922K &
  38.22 &
  0.9612 &
  33.82 &
  0.9217 &
  32.36 &
  0.9020 &
  33.04 &
  0.9365 &
  39.30 &
  0.9777 \\
4& DWConv,  $k=5, 13$ &
  915K &
  38.23 &
  0.9614 &
  33.91 &
  0.9205 &
  32.36 &
  0.9021 &
  33.07 &
  0.9364 &
  39.36 &
  0.9783 \\
5& DWConv, $k=9, 17$ &
  957K &
  38.25 &
  0.9614 &
  34.00 &
  0.9220 &
  32.34 &
  0.9019 &
  33.03 &
  0.9365 &
  39.34 &
  {\color[HTML]{9a0000} 0.9785} \\
6& DWConv, $k=15, 7$ &
  929K &
  38.23 &
  0.9614 &
  {\color[HTML]{9a0000} 34.03} &
  {\color[HTML]{9a0000} 0.9222} &
  32.35 &
  0.9022 &
  33.01 &
  0.9364 &
  39.37 &
  0.9776 \\  
7& DWDConv, $k=7, 15$&
  929K &
  38.24 &
  0.9613 &
  33.98 &
  0.9219 &
  32.35 &
  0.9021 &
  33.04 &
  0.9363 &
  39.31 &
  0.9778 \\
  8&Ours &
  929K &
  {\color[HTML]{9a0000} 38.26} &
  {\color[HTML]{9a0000} 0.9615} &
   33.99 &
   0.9221 &
  {\color[HTML]{9a0000} 32.38} &
  {\color[HTML]{9a0000} 0.9022} &
  {\color[HTML]{9a0000} 33.09} &
  {\color[HTML]{9a0000} 0.9369} &
  {\color[HTML]{9a0000} 39.42} &
  0.9778 \\
  \bottomrule
\end{tabular}%
}
\end{table*}

\begin{figure}[t]
\centering
\includegraphics[width=1\linewidth]{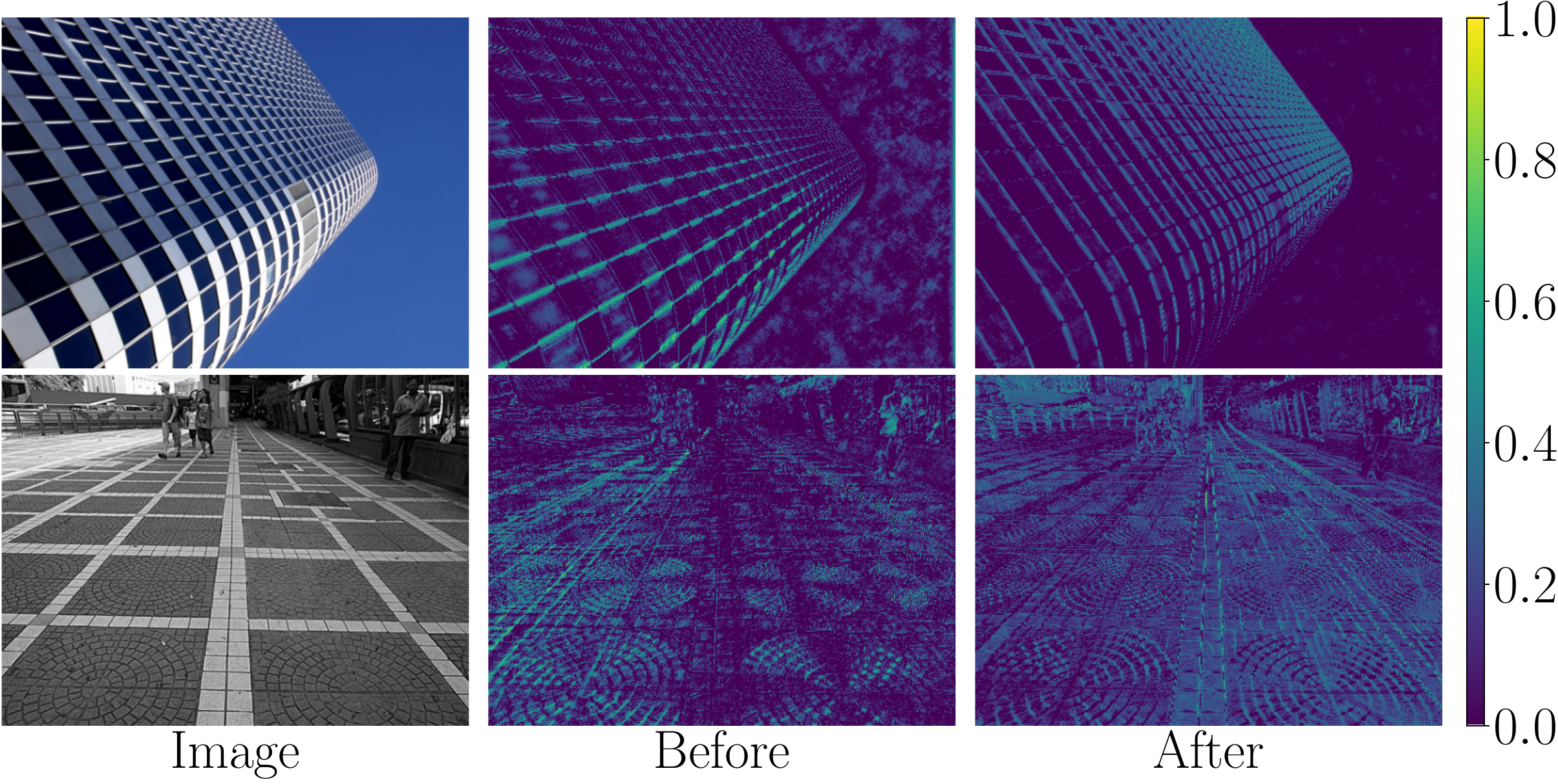}
\caption{Visualization of COFB. After processing by the COFB module, the activation maps respond more strongly to structural information in the image (such as the grid structure of the building and the squares on the ground), achieving a balance of local and global contextual features.}
\label{fig:act_map_olkb}
\end{figure}

\textbf{Effectiveness of the COFB.} Table~\ref{tab:Ab-COFB} explore the impacts of different variants of COFB. 
Compared to ours (\#7), configuration (\#1) utilized standard convolutions, mirroring the configuration in SwinIR~\cite{swinIR}; 
Despite the increased parameter count, performance decreased by 0.12dB on Urban100 and 0.16dB on Manga109, datasets that emphasize global features, while the performance decrease on Set14, which focuses more on local features, was only 0.09dB.
Subsequently, we employed only a single convolution (\#2, \#3). 
Configuration (\#2), utilizing a smaller kernel size of 7, exhibited a more pronounced performance decline on Urban100 and Manga109. 
In contrast, (\#3), employing a larger kernel size, resulted in a significant PSNR decrease of 0.17dB on Set14, compared to a 0.07dB decrease for configuration (\#2) with a kernel size of 7. 
Then, we evaluated the impact of replacing both the convolutions and their sizes (\#4, \#5), with both configurations yielding inferior results. 
This highlights the necessity and effectiveness of our cascaded overlapping convolutions for fusing multi-granularity features and demonstrates that our COFB attends to both local and global features. Furthermore, we investigated the effect of the kernel sequence order in the cascaded structure. Compared to our small-to-large configuration (\#8), the reversed large-to-small sequence (\#6) exhibited a noticeable performance degradation on structured datasets, with a drop of 0.08dB on Urban100 and 0.05dB on Manga109.
In addition, we tested replacing the depthwise convolutions with dilated depthwise convolutions (\#7). The results, consistent with our observations for the MRFE, indicated that neglecting contiguous features is detrimental to the model's performance.
Our COFB (\#8) employs cascaded large-kernel convolutions with kernel sizes of 7 and 15 to aggregate local and global contextual features, preserving the faithful details with high efficiency.

In Fig.~\ref{fig:act_map_olkb}, we analyze the feature activation maps before and after the input passes through the COFB. 
The results clearly demonstrate that the COFB effectively balances dependencies across different scales by leveraging cascaded overlapping large-kernel receptive fields, leading to noise suppression and enhanced feature clarity.

\begin{table}[t]
\caption{Ablation experiments on different feed-forward networks. CAFFN~\cite{MogaNet} was employed in our EchoSR. The best results are highlighted in {\color[HTML]{9a0000} red}.}
\label{tab:Ab-FFN}
\resizebox{1\columnwidth}{!}{%
\begin{tabular}{@{}lccccccc@{}}
\toprule
\multicolumn{1}{c}{} &
   &
  \multicolumn{2}{c}{Set14} &
  \multicolumn{2}{c}{Urban100} &
  \multicolumn{2}{c}{Manga109} \\ \cmidrule(l){3-8} 
 {\multirow{-2.5}{*}{Variant}} &
  \multirow{-2.5}{*}{\# Params} &
  PSNR &
  SSIM &
  PSNR &
  SSIM &
  PSNR &
  SSIM \\ \midrule

FFN~\cite{AttIsAll} &
  910K &
  33.91 &
  0.9210 &
  33.04 &
  0.9365 &
  39.34 &
  0.9780 \\
CFFN~\cite{PVTV2} &
  927K &
  33.97 &
  {\color[HTML]{9a0000} 0.9222} &
  33.00 &
  0.9363 &
  39.38 &
  0.9783 \\
GateFFN~\cite{SeeMore} &
  922K &
  {\color[HTML]{9a0000} 33.99} &
  0.9221 &
  33.06 &
  0.9366 &
  39.41 &
  {\color[HTML]{9a0000} 0.9786} \\ 
  CAFFN~\cite{MogaNet} & 929K & {\color[HTML]{9a0000} 33.99} & 0.9221 & {\color[HTML]{9a0000} 33.09} & {\color[HTML]{9a0000} 0.9369} & {\color[HTML]{9a0000} 39.42} & 0.9778 \\
  \bottomrule
\end{tabular}%
}
\end{table}

\begin{figure}[t]
\centering
\includegraphics[width=1\linewidth]{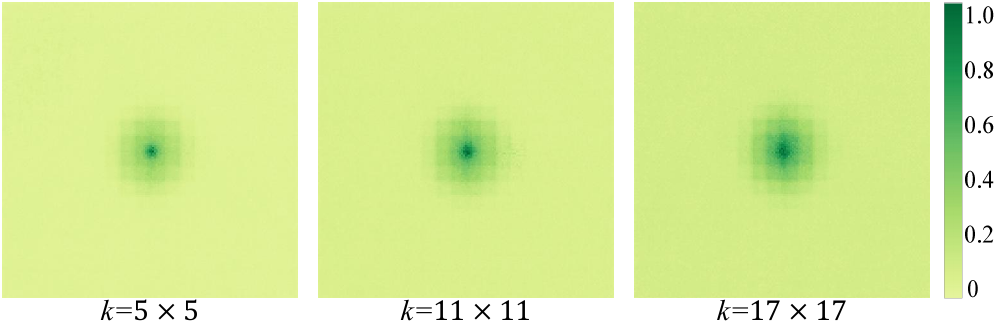} 
\caption{Visualization of the ERF for the three parallel branches in the MRFE module. A more widely discributed dark area indicates a larger ERF. The heatmaps demonstrate that the $\{5,11,17\}$ configuration captures multi-scale contextual dependencies, where larger kernels provide holistic structural coverage while smaller kernels maintain high-density local responses.}
\label{fig:MRFE_erf}
\end{figure}

\textbf{Effectiveness of the FFN.} As shown in Table~\ref{tab:Ab-FFN}, we further explore the impact of different FFNs on the performance of our EchoSR. We compared four FFN variants: traditional FFN~\cite{AttIsAll}, CFFN~\cite{InceptionNeXt}, GateFFN~\cite{SeeMore}, and CAFFN~\cite{MogaNet}. These four FFN structures are roughly equivalent in terms of parameter quantity.
We found that the EchoSR model integrated with CAFFN performed best in terms of PSNR indicator, which was improved by 0.03dB on the Urban100 dataset and 0.01dB on the Manga109 dataset compared with the second-best GateFFN. 
Although CAFFN did not show a significant advantage in SSIM, considering the trade-off of overall performance, we finally chose CAFFN as the feed-forward network module in EchoSR.

\subsection{Discussion}
\label{sec:discussion}

In this section, we discuss the underlying rationale for the multi-scale receptive field expansion (MRFE) and cross-scale overlapping fusion block (COFB) modules. Following Ding et al.'s practice~\cite{RepLKNet}, we utilized the Effective Receptive Field (ERF)~\cite{erf} to analyse the design rationale for the kernel sizes.

\textbf{Rationale for MRFE.} MRFE employs a multi-scale design strategy with kernel sizes of $5 \times 5$, $11 \times 11$, and $17 \times 17$. The selection of kernel sizes $\{5, 11, 17\}$ follows an arithmetic progression with a constant interval of 6. This equidistant size expansion is designed to capture a balanced spectral coverage of features, ranging from high-frequency local textures ($5\times5$) to mid-range structural components ($11\times11$) and long-range contexts ($17\times17$). Kernel size intervals that are too narrow fail to capture sufficient feature diversity and constrain the receptive field. Intervals that are too wide increase the parameter load but sacrificing intermediate feature representations, as validated by the ablation study in Table~\ref{tab:more_kernel}.

Fig.~\ref{fig:MRFE_erf} visualizes of the ERF for the three parallel branches in MRFE. The $17\times17$ branch exhibits a expanded receptive field compared to the $5\times5$ baseline, highlighting its capacity of capturing large-scale structural dependencies. On the other hand, the $5\times5$ kernel effectively maintains superior local activation intensity. This multi-scale design strategy establishes a complementary mechanism where the large branch grasps the global context while the small kernel preserves local integrity. Consequently, EchoSR achieves high-precision reconstruction without sacrificing holistic background coherence.

\begin{table*}[t]
\centering
\caption{Quantitative analysis on the ERF with the high-contribution area ratio for various kernel configurations of the COFB module. The area is measured as the percentage of spatial regions required to cover specific cumulative contribution score thresholds, for instance, a value of 20.7305\% at the 70\% threshold indicates that  20.7305\% of pixels account for 70\% of the total contribution scores.}
\label{tab:erf_COFB}
\resizebox{\textwidth}{!}{%
\begin{tabular}{@{}lllllllllll@{}}
\toprule
\multirow{2.5}{*}{Threshold} &
  \multicolumn{2}{c}{Kernel size = 7,15} &
  \multicolumn{2}{c}{Kernel size = 15,7} &
  \multicolumn{2}{c}{Kernel size = 7} &
  \multicolumn{2}{c}{Kernel size = 15} &
  \multicolumn{2}{c}{Kernel size = 3} \\ \cmidrule(l){2-11} 
     & Before(\%) & After(\%) & Before(\%) & After(\%) & Before(\%) & After(\%) & Before(\%) & After(\%) & Before(\%) & After(\%) \\ \midrule
10\% & 0.0430     & 0.0508    & 0.0508     & 0.0508    & 0.0195     & 0.0273    & 0.0430     & 0.0352    & 0.0195     & 0.0195    \\
20\% & 0.1562     & 0.1836    & 0.1914     & 0.2070    & 0.0938     & 0.1172    & 0.1445     & 0.1328    & 0.0859     & 0.0781    \\
30\% & 0.4609     & 0.5273    & 0.5586     & 0.5820    & 0.3477     & 0.3906    & 0.4366     & 0.4336    & 0.3047     & 0.2500    \\
50\% & 2.0742     & 2.2578    & 2.5781     & 2.4727    & 1.8828     & 1.8633    & 2.0352     & 2.1641    & 1.5625     & 1.3867    \\
70\% & 20.7305    & 22.6919   & 31.1055    & 26.9180   & 23.6133    & 21.7695   & 22.8594    & 25.1211   & 13.2812    & 11.1055   \\
90\% & 73.3281    & 73.8828   & 74.6484    & 74.9844   & 74.2070    & 73.6953   & 73.6914    & 74.4023   & 69.5430    & 68.6953   \\
95\% & 86.6523    & 86.8984   & 87.0586    & 87.4102   & 87.0547    & 86.8633   & 86.7617    & 87.2188   & 84.6094    & 84.1602   \\ \bottomrule
\end{tabular}%
}
\end{table*}

\begin{figure*}[t]
\centering
\includegraphics[width=0.8\textwidth]{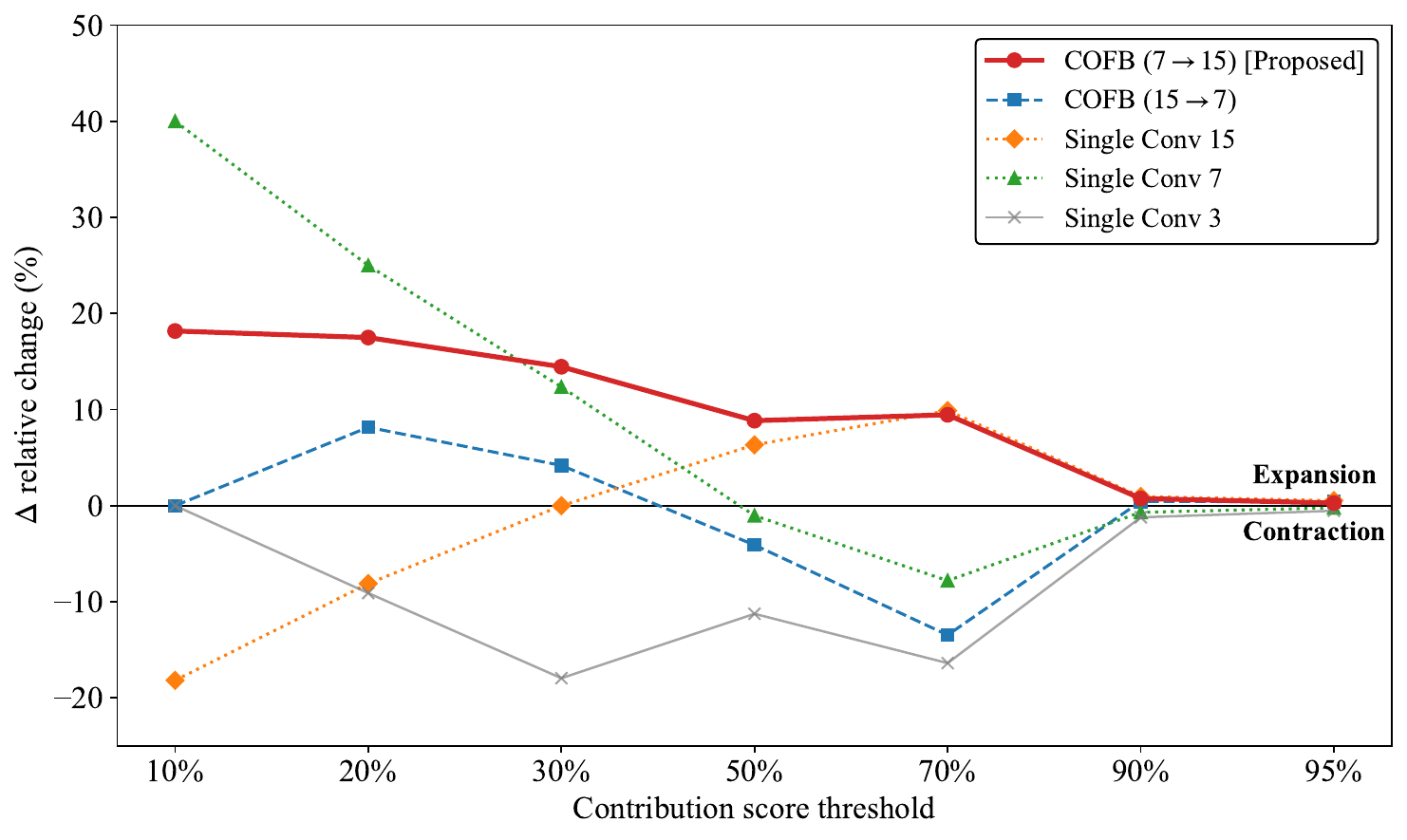}
\caption{Quantitative analysis of ERF area ratio relative change across various cumulative contribution score thresholds ($10\%$--$95\%$). The curves depict the percentage change between ``before'' and ``after'' COFB states. The proposed $7 \to 15$ cascaded sequence in COFB achieves a stable and constructive expansion of the receptive field. In contrast, the reversed $15 \to 7$ sequence and single-kernel baselines exhibit significant contraction and structural dissipation in critical core regions.}
\label{fig:erf_energy}
\end{figure*}

\textbf{Rationale for COFB.} COFB utilizes a small-to-large cascaded structure ($7 \to 15$) to prioritize local information integration prior to contextual expansion. The $7\times7$ kernel is used in the local aggregator to stabilize the signal and consolidate pixel-level information. Then, the $15\times15$ kernel propagates these refined features across a broader contextual span. As shown in Table~\ref{tab:erf_COFB}, we performed the quantitative analysis on the ERF with the high-contribution area ratio for various kernel configurations of COFB. The proposed $7 \rightarrow 15$ sequence demonstrates a consistent and constructive expansion of contribution scores . In contrast, the reversed $15 \rightarrow 7$ sequence and single-kernel variants exhibit a contraction phenomenon, particularly at the 70\% core threshold, indicating contribution score dissipation when applying large kernels to unrefined features.

Fig.~\ref{fig:erf_energy} plots the relative change of ERF area ratio between the ``before'' and ``after'' COFB states. As illustrated in Fig.~\ref{fig:erf_energy}, our proposed cascade facilitates a stable and monotonic expansion of contribution areas across critical thresholds, achieving increases of $+18.18\%$ at Th\_10\% (corresponding to an area coverage increase from 0.0430\% to 0.0508\%) and $+9.46\%$ at Th\_70\%.
In contract, alternative designs exhibit a clear contraction phenomenon. As visualized in Fig.~\ref{fig:erf_energy}, the reversed $15 \to 7$ sequence initiates with a significantly more dispersed contribution distribution (31.1055\% area coverage at Th\_70\% before processing), indicating that early large-kernel processing prematurely diffuses contribution scores into less informative regions or spatial noise. Consequently, the subsequent small kernel induces a significant contraction of $-13.46\%$ at the $70\%$ core region as it attempts to re-focus the scattered feature responses, reflecting contribution score dissipation rather than constructive refinement. This limitation is corroborated by the ablation study in Table~\ref{tab:Ab-COFB}, where the reversed $15 \to 7$ sequence yields inferior overall performance compared to the proposed $7 \to 15$ design. Similarly, a single $15\times15$ kernel exhibits an initial collapse of $-18.18\%$ at Th\_10\%. By pre-aggregating features with a smaller kernel, the subsequent large kernel of our design effectively integrates long-range context without the structural collapse caused by unrefined sparse aggregation.

At ultra-high thresholds about $90\%$, it can be observed that the area expansion becomes marginal with a change of less than $1\%$. This controlled behavior, indicated by the tail of the curves in Fig.~\ref{fig:erf_energy}, demonstrates that the module is highly selective. It strengthens the influence of core informative pixels while strictly limiting the indiscriminate propagation of distant, irrelevant background noise, thereby striking an optimal balance between long-range context and high-frequency fidelity.

\section{Conclusion}

We have presented a novel, efficient context-harnessing framework (EchoSR) for lightweight super-resolution, achieving both efficient and accurate reconstruction that surpasses prior methods in terms of both efficiency and fidelity. Our proposed context-harnessing block effectively captures deep features, with multiple concise yet effective components facilitating efficient information flow. The cross-scale overlapping fusion block achieves a delicate balance in the fusion of global and local features. Extensive experimental results demonstrate that our proposed EchoSR significantly outperforms recent lightweight and tiny state-of-the-art CNN-based, Transformer-based, and Mamba-based super-resolution methods. 

In future work, we aim to generalize our efficient context-harnessing mechanism beyond image super-resolution to other low-level vision tasks, including image denoising and deblurring. Extending this framework to capture spatio-temporal dependencies for video super-resolution is another promising direction. Concurrently, we will investigate structural re-parameterization techniques to further optimize inference latency and memory fragmentation, facilitating the deployment of high-fidelity restoration models on resource-constrained edge devices.

\bibliographystyle{elsarticle-num}

\bibliography{main}



\end{document}